\documentclass{article}

\usepackage[final]{neurips_2022}

\usepackage[utf8]{inputenc} % allow utf-8 input
\usepackage[T1]{fontenc}    % use 8-bit T1 fonts
\PassOptionsToPackage{hyphens}{url}\usepackage{hyperref}
\usepackage{url}            % simple URL typesetting
\usepackage{booktabs}       % professional-quality tables
\usepackage{amsfonts}       % blackboard math symbols
\usepackage{nicefrac}       % compact symbols for 1/2, etc.
\usepackage{microtype}      % microtypography
\usepackage{xcolor}         % colors

\usepackage{microtype}
\usepackage{graphicx}
\usepackage{subfigure}
\PassOptionsToPackage{hyphens}{url}\usepackage{hyperref}

\usepackage{amsmath}
\usepackage{amssymb}
\usepackage{mathtools}
\usepackage{amsthm}
\usepackage[capitalize,noabbrev]{cleveref}
\usepackage{caption}
\usepackage{comment}

\title{Capturing Failures of Large Language Models via Human Cognitive Biases}

\author{%
  Erik Jones \\ 
  UC Berkeley\\ 
  \texttt{erjones@berkeley.edu}\\
  \And
  Jacob Steinhardt \\ 
  UC Berkeley\\ 
  \texttt{jsteinhardt@berkeley.edu}\\
}

\newif\ifcomment
\commenttrue
\newcommand\refsec[1]{Section~\ref{sec:#1}}

\newcommand\reffig[1]{Figure~\ref{fig:#1}}

\newcommand\reftab[1]{Table~\ref{tab:#1}}
\newcommand\refapp[1]{Appendix~\ref{sec:#1}}

\ifthenelse{\isundefined{\definition}}{\newtheorem{definition}{Definition}}{}
\ifthenelse{\isundefined{\assumption}}{\newtheorem{assumption}{Assumption}}{}
\ifthenelse{\isundefined{\hypothesis}}{\newtheorem{hypothesis}{Hypothesis}}{}
\ifthenelse{\isundefined{\proposition}}{\newtheorem{proposition}{Proposition}}{}
\ifthenelse{\isundefined{\theorem}}{\newtheorem{theorem}{Theorem}}{}
\ifthenelse{\isundefined{\lemma}}{\newtheorem{lemma}{Lemma}}{}
\ifthenelse{\isundefined{\corollary}}{\newtheorem{corollary}{Corollary}}{}
\ifthenelse{\isundefined{\alg}}{\newtheorem{alg}{Algorithm}}{}
\ifthenelse{\isundefined{\example}}{\newtheorem{example}{Example}}{}
\ifcomment
\newcommand{\ej}[1]{ }
\newcommand{\js}[1]{{\bf \color{red} JS: #1}}
\else
\newcommand{\ej}[1]{}
\newcommand{\js}[1]{}
\fi
\newcommand{\python}[1]{\color{black}{\texttt{#1} }\color{black}}

\usepackage{multirow}

\renewcommand{\paragraph}{\textbf}
\begin{document}

\maketitle

\begin{abstract}
Large language models generate complex, {open-ended} outputs: instead of outputting a class label 
they write summaries, generate dialogue, or produce working code. 
In order to asses the reliability of these open-ended generation systems, 
we aim to identify qualitative categories of erroneous behavior, beyond identifying individual errors. 
To hypothesize and test for such qualitative errors, we draw inspiration from human cognitive biases---systematic patterns of deviation from rational judgement.
Specifically, we use cognitive biases as motivation to (i) generate hypotheses for problems that models may have, and
(ii) develop experiments that elicit these problems. 
Using code generation as a case study, 
we find that OpenAI’s Codex 
errs predictably based on how the input prompt is framed, adjusts outputs towards anchors, and is biased towards outputs that mimic frequent training examples. 
We then use our framework to elicit \emph{high-impact errors} such as incorrectly deleting files. 
Our results indicate that experimental methodology from cognitive science can help characterize how machine learning systems behave.\footnote{Code for this paper is available at \url{https://github.com/ejones313/codex-cog-biases}.}

\end{abstract}
\section{Introduction}
\label{sec:intro}
Recent large language models have achieved new, exciting capabilities. 
In contrast to traditional classifiers, these models can generate 
open-ended text, enabling use cases like 
summarization \citep{steinnon2020learning}, 
dialog \citep{thoppilan2022lambda}, 
and code generation \citep{chen2021codex}. 

The open-ended power of these systems, however, poses new reliability challenges. 
We must understand not only when systems err, 
but also the \emph{kinds} of errors they make, as some errors are much more costly than others. 
For example, erroneous code that does not compile is less dangerous than code that deletes all files in the home directory. 
Studying how frequently an error occurs is difficult, as the same error (e.g.~delete all files) can appear in a wide range of syntactically diverse outputs. 
In order to better reason about how complex systems err, we need methods to test whether systems make the same qualitative error across different prompts, even when the generated outputs differ. 

To study these reliability challenges, we primarily focus on {code generation} models. 
Such models complete programs from comments, descriptions of code functionality, or initial lines of code. 
Code generation is particularly amenable to study since it is \emph{objective}: generated solutions are unambiguously correct or incorrect. 
Yet it is also \emph{open-ended}: 
the set of programs a model could output is arbitrarily large, 
so the rate at which a specific program is outputted is not very descriptive. 

Many of the reliability challenges posed by code generation models, and open-ended systems broadly, also arise when studying qualitative failures in human decision making. 
These failures, called \emph{cognitive biases}, are systematic ways in which humans deviate from rational judgment \citep{tversky1974judgement}.
For example, \citeauthor{tversky1974judgement} find that humans inadequately adjust estimates away from initial values, and disproportionately recall distinctive examples.  
To uncover cognitive biases, \citeauthor{tversky1974judgement} ask questions that are crafted to systematically reveal some  qualitative irrationality. 
They uncover insights into human behavior from the diverse responses, without complete mechanistic insight into the minds that they aim to analyze. 

\begin{figure*}
    \centering
    \includegraphics[width=0.99\linewidth]{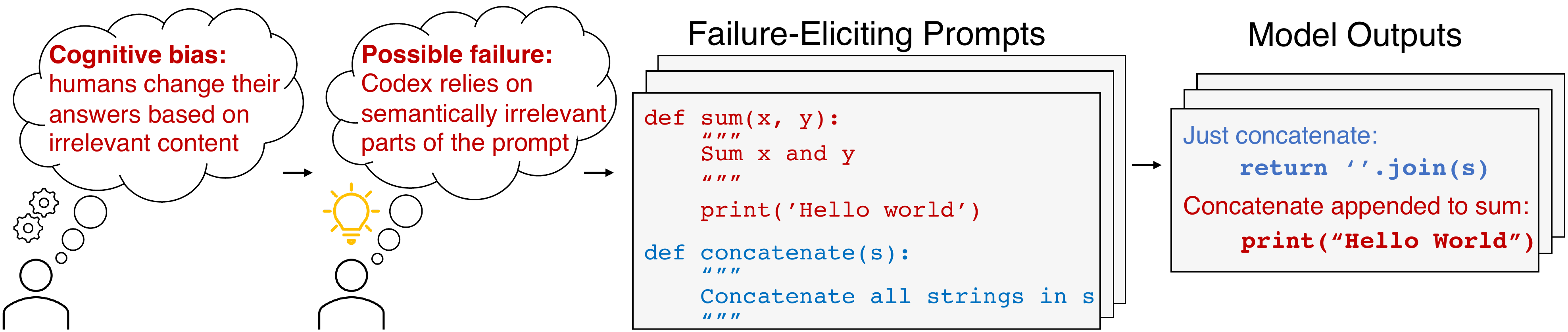}
    \caption{
    Illustration of our experimental framework. 
    We use a cognitive bias (framing effect) to inspire a potential code generation failure mode (relying on irrelevant information). 
    We then transform inputs in a way that we suspect will elicit the failure mode (prepending \texttt{sum}).
    We evaluate whether the modifications lower accuracy, and if the output is an instance of the targeted failure mode. 
    }
    \label{fig:pipeline}
\end{figure*}

In this work, we extend \citeauthor{tversky1974judgement}'s experimental methodology and results to elicit failure modes of large code and language models, without relying on complete mechanistic insight into their behavior (\reffig{pipeline}). 
Given a potential failure mode (e.g. relying on irrelevant information in the input), we construct a transformation over inputs that largely preserves semantics, but that we suspect will elicit the failure (e.g. prepending an irrelevant function). 
We first test if the model is sensitive to the transformation, by measuring if it decreases accuracy. 
Then, we check that the model outputs have elements that are indicative of the targeted failure (e.g. copies the irrelevant function). 

We draw on four different cognitive biases to hypothesize potential failures of 
OpenAI's Codex \citep{chen2021codex} and Salesforce's CodeGen \citep{nijkamp2022codegen}, 
then apply our framework to each. 
Our results indicate that these models often rely on irrelevant information when generating solutions, 
adjust solutions towards related-but-incorrect solutions, 
are biased based on training-set frequencies, 
and reverts to computationally simpler problems when faced with a complex calculation. 
We also apply our framework to OpenAI's GPT-3 \citep{brown2020gpt3}, and show that it updates its predictions towards anchors, and predictably adjusts its responses based on the question framing. 

Finally, we show that our framework can uncover 
\emph{high-impact errors}: errors that are harmful and difficult to undo. 
Specifically, we use our framework to systematically generate prompts where Codex erroneously deletes files. 
Our results indicate that experimental methodology from cognitive science can help uncover failure modes of complex machine learning systems.

\section{Related Work}
\label{sec:related_work}
\textbf{Large language models.} Recent work has developed large, capable, autoregressive language models, which predict future tokens from past tokens \citep{radford2019language, wang2021gptj, brown2020gpt3, chen2021codex, rae2021gopher}. 
These models can be used for open-ended generation tasks such as summarization \citep{steinnon2020learning, ziegler2019fine, rothe2020leveraging}, dialogue \citep{ram2018conversational, thoppilan2022lambda}, and long form question answering \citep{fan2019eli5}, among others. 
Model-generated code has been used to solve both programming and statistics questions  \citep{chen2021codex,tang2021solving}. 

There is some existing work studying failures of large language models. 
Benchmarks that measure model performance on multiple choice questions \citep{wang2019glue, wang2019superglue, hendrycks2021measuring}, 
mathematics \citep{hendrycks2021math, cobbe2021training}, long-form question answering \citep{lin2021truthful, gabriel2021go, shuster2021retrieval, krishna2021hurdles},
and coding problems \citep{hendrycks2021coding, chen2021codex} reveal inputs that the model errs on, but not the kind of error it makes. 
Another line of work shows that test-based language models can internalize bias and stereotypes \citep{sheng2019woman, nadeem2020stereoset,  groenwold2020investigating, blodgett2021stereotyping, gehman2020realtoxicityprompts}, and proposes applying fairness measurements from cognitive social sciences to machine learning systems \citep{jacobs2021measurement}. 
Some work adversarially prompts models to leak training data \citep{carlini2020extracting}, or output specific content \citep{wallace2019universal, carlini2020extracting}. 
And a final line of work identifies additional potential failures of current and future machine learning systems \citep{bender2021stochastic, bommasani2021opportunities, weidinger2021ethical}. 

\paragraph{Cognitive biases.}
\citet{tversky1974judgement} define \emph{human cognitive biases}: systematic patterns of deviation from rational judgment. 
They observe that humans employ heuristics when computing probabilities or assessing values, and that these heuristics lead to predictable errors. 
Follow-up work has added to, refined, and validated the set of known
cognitive biases \citep{tversky1973availability, tversky1981framing, strack1988priming, kahneman2002representativeness, windhager2010laying, meyer2014semantic}. 

Some known failure modes of large language models resemble cognitive biases. 
\citet{zhao2021calibrate} and \citet{liu2021what} show that the specific random samples used for few-shot learning can change GPT-3's prediction on binary and multiple choice tasks. 
Similarly, \citet{wallace2019universal} show that innocuous prompts can routinely generate toxic model output. 
Our framework builds on this work by (i) identifying the link to cognitive biases, (ii) focusing on open-ended generation, and (iii) leveraging \citeauthor{tversky1974judgement}'s experimental methodology to elicit qualitative failure modes. 

\section{Code Generation Experiments}
\label{sec:experiments}
\subsection{Models}
\label{sec:model_details}
We study two code models: OpenAI's Codex \citep{chen2021codex}, and Salesforce's CodeGen.  \citep{nijkamp2022codegen}. 
Both models are {autoregressive}---given a sequence of previous tokens, they predict the next token.
Practitioners query these code models with partial programs, docstrings, or function signatures, and obtain completions as output. 

\paragraph{Codex.} We study OpenAI's Codex, a large language model trained to generate code from docstrings \citep{chen2021codex}. 
We use the OpenAI API to query the ``davinci-001'' version of Codex, and use greedy decoding to generate solutions. 
Details of this model architecture are not public, but it is likely similar to the largest model from \citet{chen2021codex}: a 12B parameter version of GPT-3 \citep{brown2020gpt3} that is fine-tuned on GitHub instead of the CommonCrawl.

\paragraph{CodeGen.}
We additionally study the 6.2 billion parameter ``mono'' version of CodeGen, which is trained on text data and fine-tuned on GitHub. 
Unlike Codex, the weights of CodeGen are publicly available,\footnote{\url{https://github.com/salesforce/CodeGen}} so we run inference locally. 
We use greedy decoding to generate solutions. 

\begin{figure}
    \centering
    \includegraphics[width=0.99\linewidth]{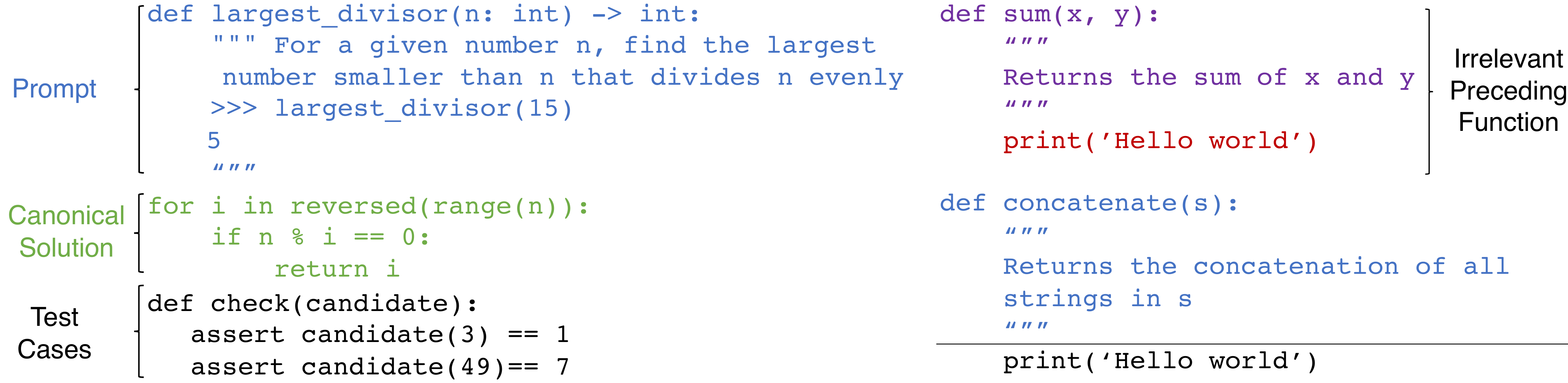}
    \caption{
    \textbf{Left.} Example of a HumanEval problem from \citet{chen2021codex}. 
    The problem contains a prompt (blue), 
    a canonical solution to the prompt (green), 
    and a few test-cases (black). 
    The prompt contains two components: 
    a function signature (first line), and a docstring (remaining lines).
    \textbf{Right.} 
    Illustration of our framing experiment. 
    The transformed prompt (everything above the black line) contains an irrelevant preceding function (IPF) prepended to a prompt from HumanEval (blue). The IPF contains a randomly chosen prompt from HumanEval (purple) and a framing line (red). The output Codex generates (below the black line) matches the framing line. 
    When we omit the random HumanEval prompt and the framing line (leaving only blue), Codex produces the correct output.
    }
    \label{fig:humaneval_framing}
\end{figure}
\subsection{Benchmarks}
In order to identify whether code models some failure mode, we need to generate prompts that elicit that failure. 
To do so, we systematically apply transformations to standard prompts. 
We use two benchmarks as sources of prompts to transform: HumanEval, and MathEquations.

\paragraph{HumanEval.} 
We use the HumanEval benchmark as a diverse source of ``normal'' prompts \citep{chen2021codex}. 
HumanEval contains 164 programming problems, each of which includes a function signature and a docstring. 
The docstring contains an English description of the desired functionality and a few example input-output pairs. 
HumanEval also contains a canonical solution for each program, which we use in \refsec{anchoring}. 
We give an example problem from HumanEval in \reffig{humaneval_framing}. 

\paragraph{MathEquations.} We also curate a set of prompts of basic arithmetic functions. For example, we prompt Codex to ``Write a function that sums the squares of its inputs'', or ``Write a function that sums its inputs called product\_plus\_five''.  
Further details are given in Sections~\ref{sec:availability_heuristic} and \ref{sec:attribute_substitution}. 

\subsection{Empirical results}
\label{sec:codex_results}
In this section, we show how cognitive biases can (i) inspire hypotheses for potential failure modes, and (ii) help us design experiments to test these hypotheses. 
Our approach has three steps. 
First, we construct a transformation over prompts that largely preserves semantics, but that we suspect will elicit a specific cognitive-bias-inspired failure mode. 
Next, we measure if code models are sensitive to the transformation, by measuring the decrease in accuracy. 
And finally, we check that the generated output has elements that are indicative of the targeted failure mode. 
Our approach mirrors the high-level methodology from \citet{tversky1974judgement}; we empirically elicit specific failure modes using targeted prompts, without complete mechanistic insight into the system that we study. 

We draw inspiration from four cognitive biases: the framing effect (\citet{tversky1981framing}; \refsec{framing_effect}), 
anchoring (\citet{tversky1974judgement}; \refsec{anchoring}), the availability heuristic (\citet{tversky1973availability}; \refsec{availability_heuristic}), 
and attribute substitution (\citet{kahneman2002representativeness}; \refsec{attribute_substitution}). 

\subsubsection{Inspiration: Framing effect}
\label{sec:framing_effect}
\begin{table}
    \centering
\begin{tabular}{lcccccc}
& & \multicolumn{2}{c}{Functional accuracy} & \multicolumn{2}{c}{Outputs framing line} \\ 
\cmidrule(lr){3-4}\cmidrule(lr){5-6}
Framing Line & Model & \textsc{Original} & \textsc{Framed} & \textsc{Original} & \textsc{Framed} \\ \midrule
\multirow{2}{*}{\texttt{raise NotImplemented}} & \textsc{Codex} & $32.9$ & $2.4$ & $1.4$ & $91.7$ \\
& \textsc{CodeGen} & $25.6$  & $1.5$ & $0.0$ &  $79.3$  \\ \midrule
\multirow{2}{*}{\texttt{pass}} & \textsc{Codex} & $32.9$ & $3.0$ & $9.7$ & $92.7$\\
& \textsc{CodeGen} &  $25.6$  & $2.1$ & $0.0$& $78.7$  \\ \midrule
\multirow{2}{*}{\texttt{assert False}} & \textsc{Codex} & $32.9$ & $3.3$ & $0.0$ & $92.7$\\
& \textsc{CodeGen} & $25.6$ & $4.2$ & $0.1$ & $72.6$  \\ \midrule  
\multirow{2}{*}{\texttt{return False}} & \textsc{Codex} & $32.9$ & $4.9$ & $ 11.5$ & $65.6$\\
& \textsc{CodeGen} & $25.6$ & $3.6$ & $0.0$ & $64.6$   \\ \midrule  
\multirow{2}{*}{\texttt{print("Hello world!")}} & \textsc{Codex} & $32.9$ & $10.6$ & $0.0$ & $62.2$ \\
& \textsc{CodeGen} & $25.6$ & $11.0$ &$0.0$ & $58.2$ \\ \midrule 
\end{tabular}
  \caption{
  Results of the framing experiments. We compare functional accuracy and the rate at which framing line is outputted over HumanEval with (framed) and without (original) irrelevant preceding functions. 
  We find that the irrelevant preceding functions lower functional accuracy across all framing lines for Codex and CodeGen. 
  Moreover, we find that the outputted function often appears verbatim in the generated output, 
  suggesting that both models rely on irrelevant information in the prompt. 
  }
  \label{tab:framing}
\end{table}
We first draw inspiration from \emph{the framing effect}: predictable shifts in human responses when the same problem is framed in different ways \citep{tversky1981framing}. 
In their study identifying the effect, \citet{tversky1981framing} 
find that subjects favor certainly saving 200 people over saving 600 with probability 1/3, yet prefer losing 600 with probability 2/3 over certainly losing 400 (even though these are equivalent). 
At its core, the framing effect shows how humans can rely on semantically irrelevant information when they make decisions. 

Using the framing effect as inspiration, we hypothesize that code generation models may generate solutions exclusively from irrelevant information in the prompt.  
To elicit this failure, we transform HumanEval prompts by prepending \emph{irrelevant preceding functions}. 
Specifically, to generate irrelevant preceding functions, we combine a random prompt from HumanEval with a \emph{framing line}. 
We test five framing lines: \python{raise NotImplementedError}, \python{pass}, \python{assert False}, \python{return False}, and \python{print("Hello world!")}. 
We first check that prepending these irrelevant preceding functions decreases functional accuracy.\footnote{Following \citet{chen2021codex}, we measure performance on HumanEval with \emph{functional accuracy}: the fraction of programs that pass all of the test cases provided at the url: \url{https://github.com/openai/human-eval.}}
Next, to test if models relied on irrelevant information in the prompt, 
we measure how much more frequently the framing line appears verbatim in the generated output. 

We report the results of our framing experiments in \reftab{framing}. 
We find that adding irrelevant preceding functions consistently lowers functional accuracy, 
by between 22.3 and 30.5 points for Codex, across the different framing lines we tested. 
Moreover, both models frequently generate the framing line: 81\% of the time for Codex and 70.7\% of time for CodeGen, compared to only 4.5\% and 0.0\% over untransformed prompts respectively. 
These results suggest that code generation models can erroneously rely on irrelevant information in the prompt in predictable ways,  
even in the extreme case when doing so contradicts the type specification in the function signature (\python{return False}).  

\subsubsection{Inspiration: Anchoring}
\label{sec:anchoring}
We next draw inspiration from \emph{anchoring}: humans' tendency to insufficiently adjust their estimates away from initial values. 
For example, \citet{tversky1974judgement} find that subjects' 
median estimate for the fraction of African countries in the UN shifts from 25\% to 45\%, based on whether they were first asked if the fraction was greater or less than 10\% and 65\%, respectively. 
Anchoring captures how humans adjust to partial information, versus irrelevant information (framing effect).% (i.e. the framing effect). 

Using anchoring as inspiration, we hypothesize that code generation models may adjust their output towards related solutions, when these solutions are included in the prompt. 
To elicit this failure, we prepend \emph{anchor functions} to prompts: functions that are similar to a valid solution for a HumanEval prompt, but contain some error. 
We first check that prepending these anchor functions decreases  functional accuracy, as in \refsec{framing_effect}. 
Next, to test if models adjust their output towards related solutions, we check that the generated solution contains elements of the anchor function.  

\begin{figure}
    \centering
    \includegraphics[width=0.99\linewidth]{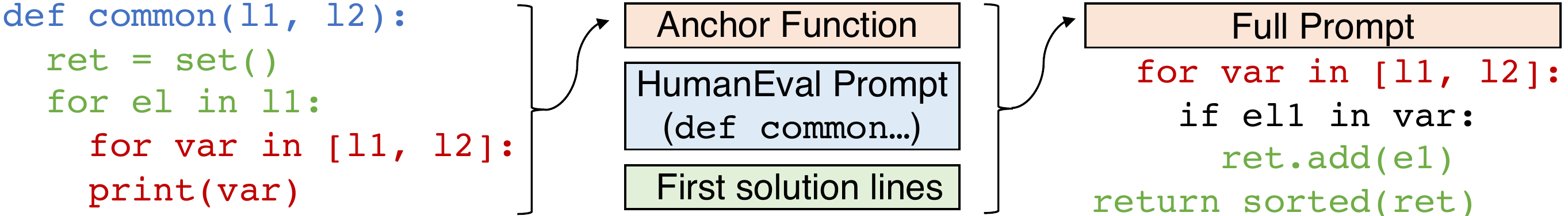}
    \caption{
    Illustration of our anchoring experiment using a real example (expanded in \reffig{anchoring_full}). We construct the anchor function (left) by taking the function signature from the HumanEval prompt (blue), appending $n$ lines of the canonical solution (green), then adding anchoring lines (red). We construct the full prompt (center) by combining the anchor function, the original HumanEval prompt, and the first $n$ lines of the canonical solution. The solution Codex generates (right) combines elements of a canonical solution (checks condition and adds to ret.), with the anchor function (for var loop). 
    }
    \label{fig:anchoring}
\end{figure}

We aim to construct anchor functions that are similar to functions in HumanEval prompts and that compile, but are incorrect. 
To do so, we take a prefix of the canonical solution, then add additional \emph{anchor lines} that produce an incorrect output. 
See \reffig{anchoring} for an example.
We describe two types of anchor lines, and how we test their influence on the generated solutions, in the following paragraphs. 

\paragraph{Print-var anchor lines.}
We first study \emph{print-var} anchor lines, 
which iterate over all variables in the function signature and print their values. 
For a function with inputs \python{var1} and \python{var2}, the associated print-var anchor lines are:  
\vspace{-1.5mm}
\begin{verbatim}
for var in [var1, var2]:
    print(var)
\end{verbatim}
\vspace{-1.5mm}
To study the influence of the print-var anchor lines on the solution, we measure how often (i) just the first line (for loop), and (ii) just the second line (print statement) appear in the generated solution. 

\begin{figure}
    \centering
    \includegraphics[width=0.99\linewidth]{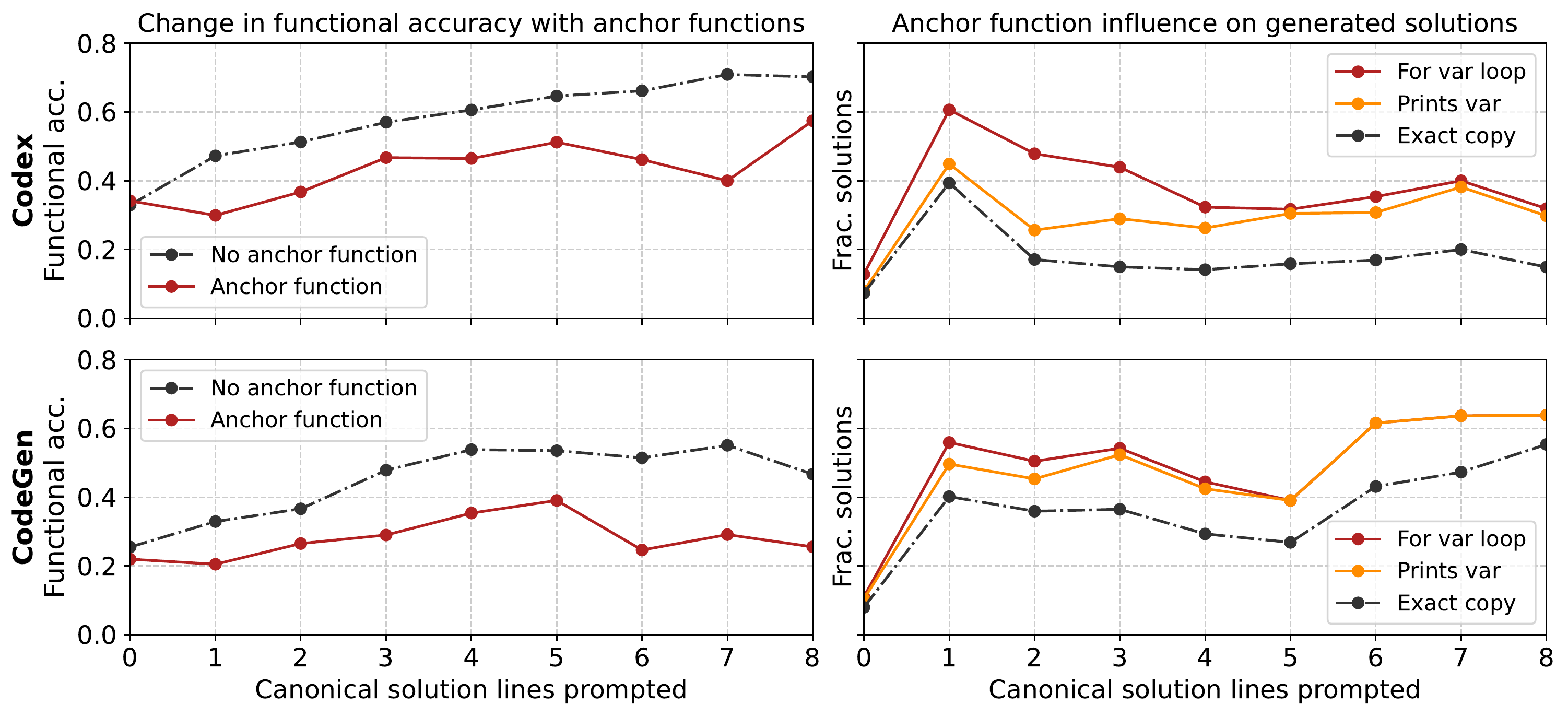}
    \caption{
    Results of the print-var anchoring experiment. 
    \textbf{Left.} We measure the functional accuracy of Codex (top) and CodeGen (bottom) with no anchor function prepended (baseline acc) and with a print-var anchor function prepended (anchor acc), and find that prepending the anchor function consistently lowers accuracy. \textbf{Right.} We measure the influence of the anchor function on the generated solution by plotting   
    the fraction of generated solutions that contain ``\python{for var in}'' from the print-var anchor prompt (for var loop), 
    the fraction of generated solutions that include ``\python{print(var)}'' (prints var),
    and the fraction of generated solutions that output the anchor function verbatim without additional content (exact copy), 
    as a function of the number of canonical solution lines added to the prompt. 
    }
    \label{fig:anchoring_printvar}
\end{figure}

\paragraph{Add-var anchor lines.}
We also study \emph{add-var} anchor lines, which return the sum of  
all variables in the function signature (converted to strings). 
For a function with inputs \python{var1} and \python{var2}, the add-var anchor lines are: 
\vspace{-1.5mm}
\begin{verbatim}
tmp = str(var1) + str(var2)
return tmp
\end{verbatim}
\vspace{-1mm}
To study the influence of the add-var anchor lines on the solution, we measure how often \python{return tmp} appears in the generated solution.

\paragraph{Print-var results.} 
In \reffig{anchoring_printvar}, we show that prepending print-var anchor functions consistently lowers Codex and CodeGens' functional accuracies across different number of prompted canonical solution lines. 
We vary the number of canonical solution lines to study prompts of different difficulties; as the number of solution lines increases, the number remaining lines models must produce decreases.\footnote{We filter out programs whose entire canonical solution would appear in the prompt; see \refapp{extra_anchoring}.}

We additionally find that elements of anchor function often appear in both models' outputs, suggesting that code generation models adjust their solutions towards related solutions. 
In \reffig{anchoring_printvar}, we see that Codex generates \python{for var} in 32\%--61\% of solutions when at least one line of the canonical solution is included, and generates \python{print(var)} in 26\%--44\% of solutions. 
CodeGen's behavior is qualitatively similar. 
Both models sometimes even incorporate the anchor lines into correct solutions; 
on Codex, the \python{for var} loop is used in a correct solution for 3\%--11\% of all outputs, while \python{print(var)} is used in a correct solution for 1\%--9\% of outputs. 

\paragraph{Control experiments.} One concern might be that models just outputs the anchor function verbatim, as in \refsec{framing_effect}, but we find that this does not explain the full results---both models include anchor lines in many solutions that do not copy the anchor function verbatim. 
We also find that changing the name of the anchor function leads to only negligible changes; see \refapp{extra_anchoring} for details. 

\paragraph{Add-var results.}
We next consider results for add-var anchor lines. 
Full results for the add-var anchor prompts are presented in \refapp{extra_anchoring} and are qualitatively similar to the print-var results.

One again, we find that prepending the anchor function consistently lowers functional accuracy. 
Moreover, the outputted solutions often include an anchor line. 
For example, Codex and CodeGen generate \python{return tmp} in 26\%--46\% and 13\%--79\% of solutions respectively, depending on how many canonical solution lines we prompt with. 
These results are not caused by models outputting the anchoring function verbatim: 
this only occurs between 7\% and 12\% of the time for Codex, and 4\% and 12\% for CodeGen. 
Overall, our findings suggest that code generation models can err by adjusting its output towards related solutions, when the solutions are included in the prompt. %, which could have ramifications for deployment. 

\subsubsection{Inspiration: Availability heuristic}
\label{sec:availability_heuristic}
We next draw inspiration from the \emph{availability heuristic}: 
the tendency of humans to evaluate how frequently an example occurs based on how easy it is to recall. 
For example, \citet{tversky1973availability} find that humans tend to incorrectly report that there are more first words that start with ``r'' and ``k'' than have third letter ``r'' and ``k'', because the former quickly come to mind.

Using the availability heuristic as motivation, we hypothesize that code generation models may err by outputting solutions to related prompts that appear more frequently in the training set.
To elicit this failure, we start with prompts that apply a unary operation before a binary operation (unary-first), then flip the order (binary-first). 
Programmers tend to apply unary operations first 
(e.g.~when computing Euclidean distances or variances), so we conjecture that they appear more frequently on GitHub. 
We first check that flipping the order of operations decreases accuracy. 
Next, to test if code generation models instead outputs related prompts that occur more frequently in the training set, we measure whether code generation models instead output the unary-first solution. 

We consider all 12 combinations of the binary operations sum, difference, and product, with unary operations square, cube, quadruple, and square root. 
Focusing on Codex,\footnote{We find that CodeGen often produces nonsensical solutions on the style of prompts used in \refsec{availability_heuristic}, \refsec{attribute_substitution}, and \refsec{risks}, so we focus primarily on Codex.} we find that accuracy drops from 50\% to 17\% when flipping the order from unary-first to binary-first. 
Among combinations where flipping the order leads to error, we find that 75\% of the binary-first outputs are the unary-first solution. 
We exhibit one such error in \reffig{availability_attribute}: 
when prompted to square the sum of its inputs, Codex generates the correct function name (\python{square\_sum}), %but instead sums the squares of its inputs.
but reverses the order of operations. 
Our results suggest that Codex can err by outputting solutions to related, frequent prompts in the training set. 

\paragraph{Control experiments.} One worry is that the dip in performance is due the instructional nature of our prompts. 
We rule this out by evaluating Codex on prompts where the docstring appears beneath the function signature and is a definition rather than command, to more closely mimic some functions on GitHub. 
We obtain qualitatively similar results on these prompts, see \refapp{extra_attribute_sub} for details. 

\subsubsection{Inspiration: Attribute substitution}
\label{sec:attribute_substitution}
Finally, we draw inspiration from \emph{attribute substitution}: 
the human tendency to respond to a complicated question using a simpler, related question \citep{kahneman2002representativeness}. 
For example, a professor when asked how likely a candidate is to be tenured, may instead respond with how impressive they found their job talk. 

\begin{figure}
    \centering
    \includegraphics[width=0.99\linewidth]{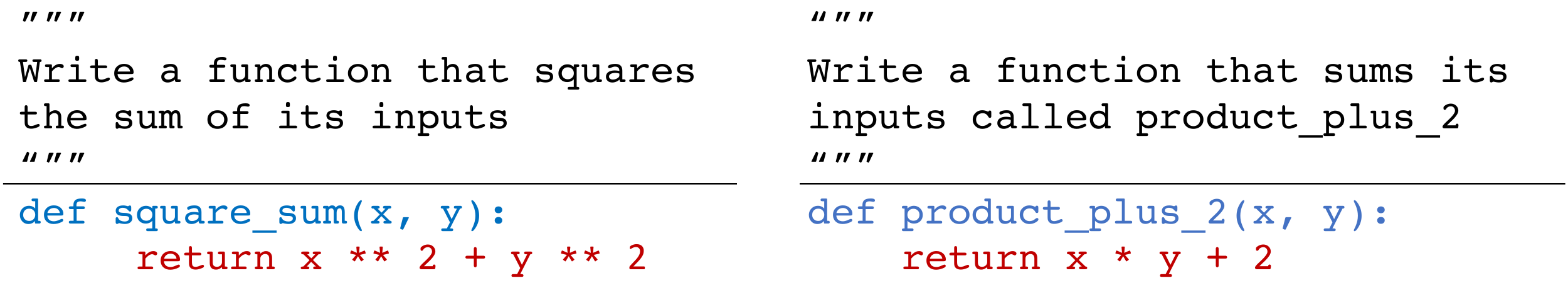}
    \caption{
    \textbf{Left.} Availability heuristic example where Codex mixes up the order of operations.
    The correct function signature (blue), \python{square\_sum} matches the prompt. 
    However, the incorrect function call (red) instead squares its inputs before summing them.
    The prompt is above the horizontal line, while the generated code is below. 
    \textbf{Right.} 
    Attribute substitution example where Codex relies on the function name to generate output. 
    Codex correctly generates the desired function name (blue), 
    but errs by using the function name instead of the prompt to generate the return statement (red). 
    }
    \label{fig:availability_attribute}
\end{figure}
Using attribute substitution as inspiration, we hypothesize that Codex may use simple-but-incorrect heuristics to generate solutions. 
To elicit this failure, we add requests for conflicting function names to MathEquation prompts.  
For example, in \reffig{availability_attribute} we prompt Codex to write a program that sums its inputs called \python{product\_plus\_2}.
We first check that adding conflicting function names decreases Codex's functional accuracy. 
Next, to test if Codex uses simple-but-incorrect heuristics to generate solutions, we check whether the generate solution matches the function name.  

We evaluate Codex using 90 MathEquation prompts where the desired solution and requested function name differ.
To construct prompts, we begin with a prompt that Codex originally solves (sum, difference, or product), then append a request for a specific, contradictory function name (see \refapp{extra_attribute_sub} for full implementation details). 

\begin{table}
  \small
  \centering
  \begin{tabular}{cccc}
    \textbf{Name location} & 
    \textbf{Correct} & \textbf{Matches function name} & \textbf{Other error} \\ 
    \cmidrule(lr){1-1}\cmidrule(lr){2-4}
    No name & 100.0 & - & 0.0 \\ 
    Docstring & 4.4 & 80.0 & 15.6 \\ 
    Function signature & 4.4 & 70.0 & 25.6 \\ 
    Name first & 4.6 & 51.7 & 43.7 \\ 
    \bottomrule\\[-2mm]
  \end{tabular}
  \caption{
    Results of the attribute substitution experiments. 
    We report accuracy when we do not request a contradictory function name (no name), 
    we request a function name in the docstring (docstring), 
    in the function signature below the docstring (function signature), or above the docstring (name first). 
    Overall, we find that Codex frequently generates solutions based on the function name. 
  }
  \label{tab:attribute_substitution}
\end{table}
We report our experimental results in \reftab{attribute_substitution}. 
When we request a conflicting function name, Codex's accuracy drops from 100\% to only 4.4\%-4.6\%.  
This finding holds whether we request the function name in the docstring, write it in the function signature below the docstring, or write the function name over a simple description on the function. 
Moreover, 
for between 52\% and 80\% of prompts, Codex responds with the function specified in the function name. 
Our results indicate that Codex can err by using simple-but-incorrect heuristics to generate solutions. 

\section{GPT-3 Results}
\label{sec:gpt3}
In this section, we extend our study from Codex to GPT-3. 
To test GPT-3 for failure modes, we try to faithfully reproduce and extend the anchoring experiment of \citet{jacowitz1995anchoring} and framing effect experiment of \citet{tversky1981framing}. 

\paragraph{Anchoring.} As in \refsec{anchoring} we study \refsec{anchoring}, we study anchoring: humans' tendency to insufficiently adjust their estimates away from an initial value \citep{tversky1974judgement}. 
We largely replicate the anchoring study presented in \citet{jacowitz1995anchoring}, but test the ``davinci-001'' version of OpenAI's GPT-3 instead of humans. 

In their original experiment, \citeauthor{jacowitz1995anchoring} asked students to estimate quantities such as the length of the Mississippi river in miles. 
They then asked new students to estimate the same quantities, but first gave them a upper or lower bound on the true answer (e.g.~the Mississippi river is longer than 700 miles), which they call \emph{anchors}. 
They find that students tend to underestimate the true quantity when prompted with the lower anchor, 
and overestimate it when prompted with the upper anchor. 

We adapt the anchoring study from \citet{jacowitz1995anchoring} by finding the true answer for 14 of their 15 original questions\footnote{We omit a question asking for the number of bars in Berkeley, CA, since the answer is ambiguous},
then computing upper and lower anchors by increasing and decreasing the true answer by a fixed percentage $p$. 
See \refapp{extra_gpt3} for a full list of questions and true answers.
As an example, if the actual answer is 2000 and $p$ is 50\%, the upper anchor is 3000 and the lower anchor is 1000. 
We use this bound as an anchor, so that a typical prompt might be: 

\vspace{-4.5mm}
\begin{align*}
    &\underbrace{\text{The length of the Mississippi river (in miles) is greater than {\color{blue} $1000$}.}}_{\textrm{anchor}} \\ 
    &\text{What is the length of the Mississippi River (in miles)? Answer:}
\end{align*}
\vspace{-4.5mm}

\begin{table}
  \small
  \centering
  \begin{tabular}{ccccc}
    &\multicolumn{4}{c}{\textbf{Anchor change to model output}}\\
    $p$ & 
    No change & Towards anchor & Away from anchor & Gibberish \\\cmidrule(lr){1-1}\cmidrule(lr){2-5}
    20\% & 10.7 & 28.6 & 10.7 & 50.0 \\ 
    50\%& 14.3 & 42.9 & 10.7 & 32.1\\
    \bottomrule\\[-2mm]
  \end{tabular}
  \caption{
  Results of the adaptation of the anchoring study from \citet{jacowitz1995anchoring} on GPT-3. We consider anchors that are 20\% and 50\% increases and decreases from the ground truth answer, and measure how often GPT-3's revised prediction does not change, shifts towards / away from the anchor, or is gibberish, aggregated across lower and upper anchors. 
}
\label{tab:gpt3_anchoring}
\end{table}

To study anchoring in GPT-3, we measure how prepending the anchor changes GPT-3's estimate. 
We categorize four potential changes: the estimate does not change, the estimate shifts towards the anchor, the estimate shifts away from the anchor, and the estimate is gibberish.
We report the results in \reftab{gpt3_anchoring} for $p \in \{20\%, 50\%\}$. 
We find that GPT-3 routinely updates its estimate when an anchor is prepended, and tends to shift the estimate towards the anchor. 
We also find that while GPT-3's updated estimate sometimes matches the anchor exactly (67\% of the time), 
it also often lands between the anchor and the original prediction, mirroring the behavior of humans. 

Our replication has a few limitations. 
Like the original study our sample size is small, we construct prompts with templates, and many of the outputs---on average 41\%---are gibberish. 
Nevertheless, our results suggest that GPT-3 incorporates the anchor during estimation. 

\paragraph{Framing effect.} 
As in \refsec{framing_effect}, we study the framing effect: predictable shifts in human responses when the same problem is framed in different ways. 
We largely replicate the framing experiment presented in \citet{tversky1981framing}: 
we compare GPT-3's responses to two equivalent decisions: choosing to either deterministically save (or let die) some fraction of a population, or to probabilistically save (let die) the whole population. 

We measure the rate at which GPT-3 chooses the probabilistic option across different population sizes and different fractions / probabilities. 
See \refsec{gpt3-framing} for full results. 
When using the probability in the original study, GPT-3 qualitatively mirrors humans: it chooses the probabilistic option far more frequently under the ``not save'' framing than under the ``save framing''. 
However, for higher probabilities, GPT-3 consistently chooses the probabilistic option for both framings; 
we conjecture that humans could exhibit similar behavior in this regime, since the probabilistic option is more certain. 
Overall, our results suggest that GPT-3 selects different options based on the framing, and could be a test-bed to identify qualitative human behaviors without running full human studies.  

\section{High-Impact Errors}
\label{sec:risks}
We have shown how our framework helps us elicit failures of large language models. 
In this section, we use our framework to construct cases where Codex makes \emph{high-impact errors}: 
harmful errors that are hard to undo. 
Specifically, we construct prompts where Codex incorrectly deletes files.

\begin{figure}
    \centering
    \includegraphics[width=0.99\linewidth]{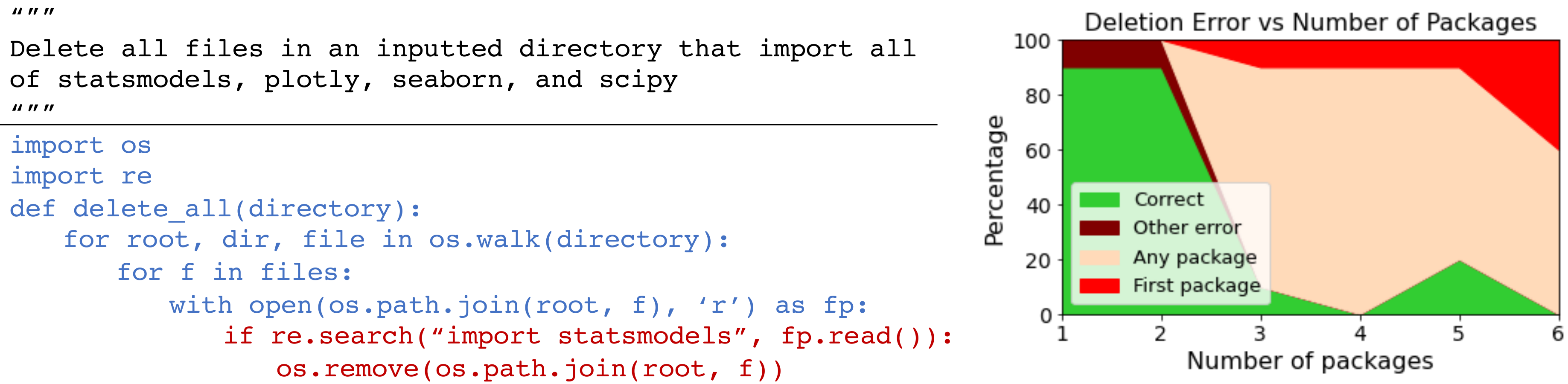}
    \caption{
    \textbf{Left.} Example where Codex incorrectly deletes files. 
    We prompt Codex to delete files containing all of statsmodels, plotly, seaborn, and scipy. 
    Codex correctly iterates through all files in the inputted directory (blue),
    but then incorrectly deletes all files containing statsmodels (red), as attribute substitution suggests. \textbf{Right.} Plot describing the errors Codex makes as a function of the number of packages. We find that Codex often incorrectly deletes files if they contain any of the listed packages, and relies more on just the first package as the number of packages increases. 
    }
    \label{fig:deletion}
\end{figure}

As in \refsec{attribute_substitution} we draw inspiration from attribute substitution: the tendency of humans to respond to a complex question with a simpler, related question. 
Using attribute substitution as motivation, we hypothesize that Codex may simplify complex expressions such as conjunctions. 
Instead of checking all components of a conjunction at once, 
it might ``give up'' and consider subsets of the components individually
(e.g.~checking for $A$ or $A \lor B$ instead of $A \land B$). 
To elicit this failure, we prompt Codex to delete files containing specific sets of package imports; see \reffig{deletion} for an example. 
We measure how often Codex generates a simpler output that erroneously deletes files, as well as how often it produces the correct output. 
See \refapp{extra_deletion} for additional details.

We test for two types of simpler outputs: code deleting all files containing first package in the set (i.e. $A$ instead of $A \land B$), and code deleting all files containing any package in the set (i.e. $A \lor B$ instead of $A \land B$). 
The latter operation is computationally simpler than checking if a file contains all packages, 
since Codex can delete a file whenever a single package in the set appears. 

In \reffig{deletion}, we illustrate the breakdown of the errors Codex makes as a function of the number of package imports in the prompt. 
We find that Codex erroneously deletes files on at least 80\% of prompts when the number of package imports is at least three, despite producing a correct output on 90\% of prompts when the number of packages is at most two. 
Moreover, we find that Codex increasingly errs by using only the first package as the problem gets more challenging (i.e. the number of packages increases), as attribute substitution predicts. 

\paragraph{Control experiments.} To very that our findings generalize to different classes of realistic prompts, we test Codex on prompts containing a descriptive docstring beneath the function signature \texttt{delete\_all\_with\_libraries(directory)}. We observe qualitatively similar results, though we find more instances of low-impact errors; see \refapp{additional_deletion_results} for details. 
Overall, our results demonstrate how our framework can preemptively elicit high-impact errors, like erroneous deletions.

\section{Discussion}
\label{sec:discussion}
In this work, we identify and test for classes of errors that open-ended generation systems can make, using cognitive biases as motivation. 
To do so, we generate hypotheses for potential qualitative failure modes, then construct transformations over prompts that elicit these failures. 
Our experiments uncover deficiencies of Codex, CodeGen, and GPT-3, and elicit high-impact errors that are challenging to undo. 
While we focus on a few specific failure modes, future work could apply our framework to uncover additional failures. 
Moreover, our framework queries systems as a black-box, so it could be used to quickly probe for errors in future systems as they are released. 

Some of our results highlight how optimizing likelihood could be at odds with human intent. For example, over GitHub, programs may more often match their function signature than docstring (\refsec{availability_heuristic}), or tend to complete to \texttt{pass} if the preceding function does (\refsec{framing_effect}). 
Nevertheless, our results elicit qualitative errors regardless of the ``correct'' behavior (i.e. even when what is incorrect and correct flips), and demonstrate the importance of documenting qualitative failures.  

The reliability challenges posed by the open-ended generation systems that we study sometimes also apply to classifiers.  
Some classification errors can be more costly than others \citep{oakden2020hidden}, 
classifiers may use irrelevant information to make predictions \citep{sagawa2020group}, 
and input-level transformations like universal adversarial triggers \citep{wallace2019universal} and distribution shifts \citep{hendrycks2019benchmarking} induce errors.
However, while classification errors may be succinctly summarized with a confusion matrix, generation errors cannot, since each output appears infrequently. 
To tame the large output space, our transformations must induce categories of errors that we can reliably measure. 
Despite this additional constraint, we are able to construct \emph{model-agnostic} transformations: we do not use the training data, model parameters, or even output logits. 
Our success in this restricted setting demonstrates the comparative brittleness of completion systems. 

We present a method to systematically elicit errors from large language models. 
While we believe our work is important to understand model behavior, bad actors could exploit the errors we reveal (e.g. by deleting files on systems with a Codex back-end). 
Nevertheless, we introduce new robustness challenges for developers and identify misuses of these models, which we feel supersedes this risk. 

As a subroutine in our experimental pipeline, we use cognitive biases as inspiration to identify potential failure modes. 
This is an example of using a reference system---a system that is analogous to the ML models we study in some meaningful way---to generate insights into ML systems \citep{steinhardt2022anchor}. 
We use humans as the reference, focusing specifically on their susceptibility to cognitive biases. 
Other references, such as complex systems or evolution, may uncover new errors and insights. 
Moreover, ML systems could additionally err in ways that known systems do not, so it will also be useful to have \emph{intrinsic} methods for characterizing model errors. 
Overall, our work underscores the need for more extensive testing of generative ML systems before their widespread deployment. 

\section*{Acknowledgements}
We thank the anonymous reviewers, Ruiqi Zhong, Jean-Stanislas Denain, Aditi Raghunathan, Jessy Lin, and Lawrence Chan for feedback. 
This work was supported by NSF Award Grant no. 1804794. 

\bibliography{all.bib}

\begin{thebibliography}{48}
\providecommand{\natexlab}[1]{#1}
\providecommand{\url}[1]{\texttt{#1}}
\expandafter\ifx\csname urlstyle\endcsname\relax
  \providecommand{\doi}[1]{doi: #1}\else
  \providecommand{\doi}{doi: \begingroup \urlstyle{rm}\Url}\fi

\bibitem[Bender et~al.(2021)Bender, Gebru, McMillan-Major, and
  Shmitchel]{bender2021stochastic}
Emily Bender, Timnit Gebru, Angelina McMillan-Major, and Shmargaret Shmitchel.
\newblock On the dangers of stochastic parrots: Can language models be too big?
\newblock In \emph{ACM Conference on Fairness, Accountability, and Transparency
  (FAccT)}, 2021.

\bibitem[Blodgett et~al.(2021)Blodgett, Lopez, Olteanu, Sim, and
  Wallach]{blodgett2021stereotyping}
Su~Lin Blodgett, Gilsinia Lopez, Alexandra Olteanu, Robert Sim, and Hanna
  Wallach.
\newblock Stereotyping norwegian salmon: An inventory of pitfalls in fairness
  benchmark datasets.
\newblock In \emph{Association for Computational Linguistics (ACL)}, 2021.

\bibitem[Bommasani et~al.(2021)Bommasani, Hudson, Adeli, Altman, Arora, von
  Arx, Bernstein, Bohg, Bosselut, Brunskill, Brynjolfsson, Buch, Card,
  Castellon, Chatterji, Chen, Creel, Davis, Demszky, Donahue, Doumbouya,
  Durmus, Ermon, Etchemendy, Ethayarajh, Fei-Fei, Finn, Gale, Gillespie, Goel,
  Goodman, Grossman, Guha, Hashimoto, Henderson, Hewitt, Ho, Hong, Hsu, Huang,
  Icard, Jain, Jurafsky, Kalluri, Karamcheti, Keeling, Khani, Khattab, Koh,
  Krass, Krishna, Kuditipudi, Kumar, Ladhak, Lee, Lee, Leskovec, Levent, Li,
  Li, Ma, Malik, Manning, Mirchandani, Mitchell, Munyikwa, Nair, Narayan,
  Narayanan, Newman, Nie, Niebles, Nilforoshan, Nyarko, Ogut, Orr,
  Papadimitriou, Park, Piech, Portelance, Potts, Raghunathan, Reich, Ren, Rong,
  Roohani, Ruiz, Ryan, Ré, Sadigh, Sagawa, Santhanam, Shih, Srinivasan,
  Tamkin, Taori, Thomas, Tramèr, Wang, Wang, Wu, Wu, Wu, Xie, Yasunaga, You,
  Zaharia, Zhang, Zhang, Zhang, Zhang, Zheng, Zhou, and
  Liang]{bommasani2021opportunities}
Rishi Bommasani, Drew~A. Hudson, Ehsan Adeli, Russ Altman, Simran Arora, Sydney
  von Arx, Michael~S. Bernstein, Jeannette Bohg, Antoine Bosselut, Emma
  Brunskill, Erik Brynjolfsson, Shyamal Buch, Dallas Card, Rodrigo Castellon,
  Niladri Chatterji, Annie Chen, Kathleen Creel, Jared~Quincy Davis, Dorottya
  Demszky, Chris Donahue, Moussa Doumbouya, Esin Durmus, Stefano Ermon, John
  Etchemendy, Kawin Ethayarajh, Li~Fei-Fei, Chelsea Finn, Trevor Gale, Lauren
  Gillespie, Karan Goel, Noah Goodman, Shelby Grossman, Neel Guha, Tatsunori
  Hashimoto, Peter Henderson, John Hewitt, Daniel~E. Ho, Jenny Hong, Kyle Hsu,
  Jing Huang, Thomas Icard, Saahil Jain, Dan Jurafsky, Pratyusha Kalluri,
  Siddharth Karamcheti, Geoff Keeling, Fereshte Khani, Omar Khattab, Pang~Wei
  Koh, Mark Krass, Ranjay Krishna, Rohith Kuditipudi, Ananya Kumar, Faisal
  Ladhak, Mina Lee, Tony Lee, Jure Leskovec, Isabelle Levent, Xiang~Lisa Li,
  Xuechen Li, Tengyu Ma, Ali Malik, Christopher~D. Manning, Suvir Mirchandani,
  Eric Mitchell, Zanele Munyikwa, Suraj Nair, Avanika Narayan, Deepak
  Narayanan, Ben Newman, Allen Nie, Juan~Carlos Niebles, Hamed Nilforoshan,
  Julian Nyarko, Giray Ogut, Laurel Orr, Isabel Papadimitriou, Joon~Sung Park,
  Chris Piech, Eva Portelance, Christopher Potts, Aditi Raghunathan, Rob Reich,
  Hongyu Ren, Frieda Rong, Yusuf Roohani, Camilo Ruiz, Jack Ryan, Christopher
  Ré, Dorsa Sadigh, Shiori Sagawa, Keshav Santhanam, Andy Shih, Krishnan
  Srinivasan, Alex Tamkin, Rohan Taori, Armin~W. Thomas, Florian Tramèr,
  Rose~E. Wang, William Wang, Bohan Wu, Jiajun Wu, Yuhuai Wu, Sang~Michael Xie,
  Michihiro Yasunaga, Jiaxuan You, Matei Zaharia, Michael Zhang, Tianyi Zhang,
  Xikun Zhang, Yuhui Zhang, Lucia Zheng, Kaitlyn Zhou, and Percy Liang.
\newblock On the opportunities and risks of foundation models.
\newblock \emph{arXiv preprint arXiv:2108.07258}, 2021.

\bibitem[Brown et~al.(2020)Brown, Mann, Ryder, Subbiah, Kaplan, Dhariwal,
  Neelakantan, Shyam, Sastry, Askell, Agarwal, Herbert-Voss, Krueger, Henighan,
  Child, Ramesh, Ziegler, Wu, Winter, Hesse, Chen, Sigler, Litwin, Gray, Chess,
  Clark, Berner, McCandlish, Radford, Sutskever, and Amodei]{brown2020gpt3}
Tom~B. Brown, Benjamin Mann, Nick Ryder, Melanie Subbiah, Jared Kaplan,
  Prafulla Dhariwal, Arvind Neelakantan, Pranav Shyam, Girish Sastry, Amanda
  Askell, Sandhini Agarwal, Ariel Herbert-Voss, Gretchen Krueger, Tom Henighan,
  Rewon Child, Aditya Ramesh, Daniel~M. Ziegler, Jeffrey Wu, Clemens Winter,
  Christopher Hesse, Mark Chen, Eric Sigler, Mateusz Litwin, Scott Gray,
  Benjamin Chess, Jack Clark, Christopher Berner, Sam McCandlish, Alec Radford,
  Ilya Sutskever, and Dario Amodei.
\newblock Language models are few-shot learners.
\newblock \emph{arXiv preprint arXiv:2005.14165}, 2020.

\bibitem[Carlini et~al.(2020)Carlini, Tramer, Wallace, Jagielski, Herbert-Voss,
  Lee, Roberts, Brown, Song, Erlingsson, Oprea, and
  Raffel]{carlini2020extracting}
Nicholas Carlini, Florian Tramer, Eric Wallace, Matthew Jagielski, Ariel
  Herbert-Voss, Katherine Lee, Adam Roberts, Tom Brown, Dawn Song, Ulfar
  Erlingsson, Alina Oprea, and Colin Raffel.
\newblock Extracting training data from large language models.
\newblock \emph{arXiv preprint arXiv:2012.07805}, 2020.

\bibitem[Chen et~al.(2021)Chen, Tworek, Jun, Yuan, de~Oliveira~Pinto, Kaplan,
  Edwards, Burda, Joseph, Brockman, Ray, Puri, Krueger, Petrov, Khlaaf, Sastry,
  Mishkin, Chan, Gray, Ryder, Pavlov, Power, Kaiser, Bavarian, Winter, Tillet,
  Such, Cummings, Plappert, Chantzis, Barnes, Herbert-Voss, Guss, Nichol,
  Paino, Tezak, Tang, Babuschkin, Balaji, Jain, Saunders, Hesse, Carr, Leike,
  Achiam, Misra, Morikawa, Radford, Knight, Brundage, Murati, Mayer, Welinder,
  McGrew, Amodei, McCandlish, Sutskever, and Zaremba]{chen2021codex}
Mark Chen, Jerry Tworek, Heewoo Jun, Qiming Yuan, Henrique~Ponde
  de~Oliveira~Pinto, Jared Kaplan, Harri Edwards, Yuri Burda, Nicholas Joseph,
  Greg Brockman, Alex Ray, Raul Puri, Gretchen Krueger, Michael Petrov, Heidy
  Khlaaf, Girish Sastry, Pamela Mishkin, Brooke Chan, Scott Gray, Nick Ryder,
  Mikhail Pavlov, Alethea Power, Lukasz Kaiser, Mohammad Bavarian, Clemens
  Winter, Philippe Tillet, Felipe~Petroski Such, Dave Cummings, Matthias
  Plappert, Fotios Chantzis, Elizabeth Barnes, Ariel Herbert-Voss,
  William~Hebgen Guss, Alex Nichol, Alex Paino, Nikolas Tezak, Jie Tang, Igor
  Babuschkin, Suchir Balaji, Shantanu Jain, William Saunders, Christopher
  Hesse, Andrew~N. Carr, Jan Leike, Josh Achiam, Vedant Misra, Evan Morikawa,
  Alec Radford, Matthew Knight, Miles Brundage, Mira Murati, Katie Mayer, Peter
  Welinder, Bob McGrew, Dario Amodei, Sam McCandlish, Ilya Sutskever, and
  Wojciech Zaremba.
\newblock Evaluating large language models trained on code.
\newblock \emph{arXiv preprint arXiv:2107.03374}, 2021.

\bibitem[Cobbe et~al.(2021)Cobbe, Kosaraju, Bavarian, Chen, Jun, Kaiser,
  Plappert, Tworek, Hilton, Nakano, Hesse, and Schulman]{cobbe2021training}
Karl Cobbe, Vineet Kosaraju, Mohammad Bavarian, Mark Chen, Heewoo Jun, Lukasz
  Kaiser, Matthias Plappert, Jerry Tworek, Jacob Hilton, Reiichiro Nakano,
  Christopher Hesse, and John Schulman.
\newblock Training verifiers to solve math word problems.
\newblock \emph{arXiv preprint arXiv:2110.14168}, 2021.

\bibitem[Fan et~al.(2019)Fan, Jernite, Perez, Grangier, Weston, and
  Auli]{fan2019eli5}
Angela Fan, Yacine Jernite, Ethan Perez, David Grangier, Jason Weston, and
  Michael Auli.
\newblock {ELI5}: Long form question answering.
\newblock In \emph{Association for Computational Linguistics (ACL)}, 2019.

\bibitem[Gabriel et~al.(2021)Gabriel, Celikyilmaz, Jha, Choi, and
  Gao]{gabriel2021go}
Saadia Gabriel, Asli Celikyilmaz, Rahul Jha, Yejin Choi, and Jianfeng Gao.
\newblock {GO FIGURE}: A meta evaluation of factuality in summarization.
\newblock In \emph{Findings of the Association for Computational Linguistics
  (Findings of ACL)}, 2021.

\bibitem[Gehman et~al.(2020)Gehman, Gururangan, Sap, Choi, and
  Smith]{gehman2020realtoxicityprompts}
Samuel Gehman, Suchin Gururangan, Maarten Sap, Yejin Choi, and Noah~A Smith.
\newblock Realtoxicityprompts: Evaluating neural toxic degeneration in language
  models.
\newblock \emph{arXiv preprint arXiv:2009.11462}, 2020.

\bibitem[Groenwold et~al.(2020)Groenwold, Ou, Parekh, Honnavalli, Levy, Mirza,
  and Wang]{groenwold2020investigating}
Sophie Groenwold, Lily Ou, Aesha Parekh, Samhita Honnavalli, Sharon Levy, Diba
  Mirza, and William~Yang Wang.
\newblock Investigating african-american vernacular english in
  transformer-based text generation.
\newblock In \emph{Empirical Methods in Natural Language Processing (EMNLP)},
  2020.

\bibitem[Hendrycks and Dietterich(2019)]{hendrycks2019benchmarking}
Dan Hendrycks and Thomas Dietterich.
\newblock Benchmarking neural network robustness to common corruptions and
  perturbations.
\newblock In \emph{International Conference on Learning Representations
  (ICLR)}, 2019.

\bibitem[Hendrycks et~al.(2021{\natexlab{a}})Hendrycks, Basart, Kadavath,
  Mazeika, Arora, Guo, Burns, Puranik, He, Song, and
  Steinhardt]{hendrycks2021coding}
Dan Hendrycks, Steven Basart, Saurav Kadavath, Mantas Mazeika, Akul Arora,
  Ethan Guo, Collin Burns, Samir Puranik, Horace He, Dawn Song, and Jacob
  Steinhardt.
\newblock Measuring coding challenge competence with {APPS}.
\newblock In \emph{Advances in Neural Information Processing Systems
  (NeurIPS)}, 2021{\natexlab{a}}.

\bibitem[Hendrycks et~al.(2021{\natexlab{b}})Hendrycks, Burns, Basart, Zou,
  Mazeika, Song, and Steinhardt]{hendrycks2021measuring}
Dan Hendrycks, Collin Burns, Steven Basart, Andy Zou, Mantas Mazeika, Dawn
  Song, and Jacob Steinhardt.
\newblock Measuring massive multitask language understanding.
\newblock In \emph{International Conference on Learning Representations
  (ICLR)}, 2021{\natexlab{b}}.

\bibitem[Hendrycks et~al.(2021{\natexlab{c}})Hendrycks, Burns, Kadavath, Arora,
  Basart, Tang, Song, and Steinhardt]{hendrycks2021math}
Dan Hendrycks, Collin Burns, Saurav Kadavath, Akul Arora, Steven Basart, Eric
  Tang, Dawn Song, and Jacob Steinhardt.
\newblock Measuring mathematical problem solving with the {MATH} dataset.
\newblock In \emph{Advances in Neural Information Processing Systems
  (NeurIPS)}, 2021{\natexlab{c}}.

\bibitem[Jacobs and Wallach(2021)]{jacobs2021measurement}
Abigail~Z. Jacobs and Hanna Wallach.
\newblock Measurement and fairness.
\newblock In \emph{ACM Conference on Fairness, Accountability, and Transparency
  (FAccT)}, 2021.

\bibitem[Jacowitz and Kahneman(1995)]{jacowitz1995anchoring}
Karen~E. Jacowitz and Daniel Kahneman.
\newblock Measures of anchoring in estimation tasks.
\newblock \emph{Personality and Social Psychology Bulletin}, 21\penalty0
  (11):\penalty0 1161--1166, 1995.

\bibitem[Kahneman and Frederick(2002)]{kahneman2002representativeness}
Daniel Kahneman and Shane Frederick.
\newblock Representativeness revisited: Attribute substitution in intuitive
  judgment.
\newblock In \emph{Heuristics and Biases: The Psychology of Intuitive
  Judgement}, pages 49--81. 2002.

\bibitem[Krishna et~al.(2021)Krishna, Roy, and Iyyer]{krishna2021hurdles}
Kalpesh Krishna, Aurko Roy, and Mohit Iyyer.
\newblock Hurdles to progress in long-form question answering.
\newblock In \emph{North American Association for Computational Linguistics
  (NAACL)}, 2021.

\bibitem[Lin et~al.(2021)Lin, Hilton, and Evans]{lin2021truthful}
Stephanie Lin, Jacob Hilton, and Owain Evans.
\newblock Truthfulqa: Measuring how models mimic human falsehoods.
\newblock \emph{arXiv preprint arXiv:2109.07958}, 2021.

\bibitem[Liu et~al.(2021)Liu, Shen, Zhang, Dolan, Carin, and Chen]{liu2021what}
Jiachang Liu, Dinghan Shen, Yizhe Zhang, Bill Dolan, Lawrence Carin, and Weizhu
  Chen.
\newblock What makes good in-context examples for {GPT}-3.
\newblock \emph{arXiv preprint arXiv:2101.06804}, 2021.

\bibitem[Meyer(2014)]{meyer2014semantic}
David~E. Meyer.
\newblock Semantic priming well established.
\newblock \emph{Science}, 345\penalty0 (6196):\penalty0 523--523, 2014.

\bibitem[Nadeem et~al.(2020)Nadeem, Bethke, and Reddy]{nadeem2020stereoset}
Moin Nadeem, Anna Bethke, and Siva Reddy.
\newblock Stereoset: Measuring stereotypical bias in pretrained language
  models.
\newblock \emph{arXiv preprint arXiv:2004.09456}, 2020.

\bibitem[Nijkamp et~al.(2022)Nijkamp, Pang, Hayashi, Tu, Wang, Zhou, Savarese,
  and Xiong]{nijkamp2022codegen}
Erik Nijkamp, Bo~Pang, Hiroaki Hayashi, Lifu Tu, Huam Wang, Yingbo Zhou, Silvio
  Savarese, and Caiming Xiong.
\newblock A conversational paradigm for program synthesis.
\newblock \emph{arXiv preprint arXiv:2203.13474}, 2022.

\bibitem[Oakden-Rayner et~al.(2020)Oakden-Rayner, Dunnmon, Carneiro, and
  R{\'e}]{oakden2020hidden}
Luke Oakden-Rayner, Jared Dunnmon, Gustavo Carneiro, and Christopher R{\'e}.
\newblock Hidden stratification causes clinically meaningful failures in
  machine learning for medical imaging.
\newblock In \emph{Proceedings of the ACM Conference on Health, Inference, and
  Learning}, pages 151--159, 2020.

\bibitem[Radford et~al.(2019)Radford, Wu, Child, Luan, Amodei, and
  Sutskever]{radford2019language}
Alec Radford, Jeffrey Wu, Rewon Child, David Luan, Dario Amodei, and Ilya
  Sutskever.
\newblock Language models are unsupervised multitask learners.
\newblock \emph{OpenAI Blog}, 1\penalty0 (8), 2019.

\bibitem[Rae et~al.(2021)Rae, Borgeaud, Cai, Millican, Hoffmann, Song,
  Aslanides, Henderson, Ring, Young, Rutherford, Hennigan, Menick, Cassirer,
  Powell, Driessche, Hendricks, Rauh, Huang, Glaese, Welbl, Dathathri, Huang,
  Uesato, Mellor, Higgins, Creswell, McAleese, Wu, Elsen, Jayakumar,
  Buchatskaya, Budden, Sutherland, Simonyan, Paganini, Sifre, Martens, Li,
  Kuncoro, Nematzadeh, Gribovskaya, Donato, Lazaridou, Mensch, Lespiau,
  Tsimpoukelli, Grigorev, Fritz, Sottiaux, Pajarskas, Pohlen, Gong, Toyama,
  de~Masson~d'Autume, Li, Terzi, Mikulik, Babuschkin, Clark, de~Las~Casas, Guy,
  Jones, Bradbury, Johnson, Hechtman, Weidinger, Gabriel, Isaac, Lockhart,
  Osindero, Rimell, Dyer, Vinyals, Ayoub, Stanway, Bennett, Hassabis,
  Kavukcuoglu, and Irving]{rae2021gopher}
Jack~W. Rae, Sebastian Borgeaud, Trevor Cai, Katie Millican, Jordan Hoffmann,
  Francis Song, J.~Aslanides, Sarah Henderson, Roman Ring, Susannah Young,
  Eliza Rutherford, Tom Hennigan, Jacob Menick, Albin Cassirer, Richard Powell,
  G.~V.~D. Driessche, Lisa~Anne Hendricks, Maribeth Rauh, Po-Sen Huang, Amelia
  Glaese, Johannes Welbl, Sumanth Dathathri, Saffron Huang, Jonathan Uesato,
  John F.~J. Mellor, I.~Higgins, Antonia Creswell, Nathan McAleese, Amy Wu,
  Erich Elsen, Siddhant~M. Jayakumar, Elena Buchatskaya, D.~Budden, Esme
  Sutherland, K.~Simonyan, Michela Paganini, L.~Sifre, Lena Martens,
  Xiang~Lorraine Li, A.~Kuncoro, Aida Nematzadeh, E.~Gribovskaya, Domenic
  Donato, Angeliki Lazaridou, A.~Mensch, J.~Lespiau, Maria Tsimpoukelli,
  N.~Grigorev, Doug Fritz, Thibault Sottiaux, Mantas Pajarskas, Tobias Pohlen,
  Zhitao Gong, Daniel Toyama, Cyprien de~Masson~d'Autume, Yujia Li, Tayfun
  Terzi, Vladimir Mikulik, I.~Babuschkin, Aidan Clark, Diego de~Las~Casas,
  Aurelia Guy, Chris Jones, James Bradbury, Matthew Johnson, Blake~A. Hechtman,
  Laura Weidinger, Iason Gabriel, William~S. Isaac, Edward Lockhart, Simon
  Osindero, Laura Rimell, Chris Dyer, Oriol Vinyals, Kareem~W. Ayoub, Jeff
  Stanway, L.~Bennett, D.~Hassabis, K.~Kavukcuoglu, and Geoffrey Irving.
\newblock Scaling language models: Methods, analysis \& insights from training
  gopher.
\newblock \emph{arXiv}, 2021.

\bibitem[Ram et~al.(2018)Ram, Prasad, Khatri, Venkatesh, Gabriel, Liu, Nunn,
  Hedayatnia, Cheng, Nagar, King, Bland, Wartick, Pan, Song, Jayadevan, Hwang,
  and Pettigrue]{ram2018conversational}
Ashwin Ram, Rohit Prasad, Chandra Khatri, Anu Venkatesh, Raefer Gabriel, Qing
  Liu, Jeff Nunn, Behnam Hedayatnia, Ming Cheng, Ashish Nagar, Eric King, Kate
  Bland, Amanda Wartick, Yi~Pan, Han Song, Sk~Jayadevan, Gene Hwang, and Art
  Pettigrue.
\newblock Conversational ai: The science behind the alexa prize.
\newblock \emph{arXiv preprint arXiv:1801.03604}, 2018.

\bibitem[Rothe et~al.(2020)Rothe, Narayan, and Severyn]{rothe2020leveraging}
Sascha Rothe, Shashi Narayan, and Aliaksei Severyn.
\newblock Leveraging pre-trained checkpoints for sequence generation tasks.
\newblock \emph{Transactions of the Association for Computational Linguistics
  (TACL)}, 8:\penalty0 264--280, 2020.

\bibitem[Sagawa et~al.(2020)Sagawa, Koh, Hashimoto, and Liang]{sagawa2020group}
Shiori Sagawa, Pang~Wei Koh, Tatsunori~B. Hashimoto, and Percy Liang.
\newblock Distributionally robust neural networks for group shifts: On the
  importance of regularization for worst-case generalization.
\newblock In \emph{International Conference on Learning Representations
  (ICLR)}, 2020.

\bibitem[Sheng et~al.(2019)Sheng, Chang, Natarajan, and Peng]{sheng2019woman}
Emily Sheng, Kai-Wei Chang, Premkumar Natarajan, and Nanyun Peng.
\newblock The woman worked as a babysitter: On biases in language generation.
\newblock In \emph{Empirical Methods in Natural Language Processing (EMNLP)},
  2019.

\bibitem[Shuster et~al.(2021)Shuster, Poff, Chen, Kiela, and
  Weston]{shuster2021retrieval}
Kurt Shuster, Spencer Poff, Moya Chen, Douwe Kiela, and Jason Weston.
\newblock Retrieval augmentation reduces hallucination in conversation.
\newblock \emph{arXiv preprint arXiv:2104.07567}, 2021.

\bibitem[Steinhardt(2022)]{steinhardt2022anchor}
Jacob Steinhardt.
\newblock Anchor weights for {ML}.
\newblock \emph{Bounded Regret}, 2022.

\bibitem[Stiennon et~al.(2020)Stiennon, Ouyang, Wu, Ziegler, Lowe, Voss,
  Radford, Amodei, and Christiano]{steinnon2020learning}
Nisan Stiennon, Long Ouyang, Jeff Wu, Daniel~M. Ziegler, Ryan Lowe, Chelsea
  Voss, Alec Radford, Dario Amodei, and Paul Christiano.
\newblock Learning to summarize from human feedback.
\newblock In \emph{Advances in Neural Information Processing Systems
  (NeurIPS)}, 2020.

\bibitem[Strack et~al.(1988)Strack, Martin, and Schwarz]{strack1988priming}
Fritz Strack, Leonard~L. Martin, and Nobert Schwarz.
\newblock Priming and communication: Social determinants of information use in
  judgments of life satisfaction.
\newblock \emph{European Journal of Social Psychology}, 18\penalty0
  (5):\penalty0 429--442, 1988.

\bibitem[Tang et~al.(2021)Tang, Ke, Singh, Verma, and Drori]{tang2021solving}
Leonard Tang, Elizabeth Ke, Nikhil Singh, Nakul Verma, and Iddo Drori.
\newblock Solving probability and statistics problems by program synthesis.
\newblock \emph{arXiv preprint arXiv:2111.08276}, 2021.

\bibitem[Thoppilan et~al.(2022)Thoppilan, Freitas, Hall, Shazeer, Kulshreshtha,
  Cheng, Jin, Bos, Baker, Du, Li, Lee, Zheng, Ghafouri, Menegali, Huang,
  Krikun, Lepikhin, Qin, Chen, Xu, Chen, Roberts, Bosma, Zhou, Chang, Krivokon,
  Rusch, Pickett, Meier-Hellstern, Morris, Doshi, Santos, Duke, Soraker,
  Zevenbergen, Prabhakaran, Diaz, Hutchinson, Olson, Molina, Hoffman-John, Lee,
  Aroyo, Rajakumar, Butryna, Lamm, Kuzmina, Fenton, Cohen, Bernstein, Kurzweil,
  Aguera-Arcas, Cui, Croak, Chi, and Le]{thoppilan2022lambda}
Romal Thoppilan, Daniel~De Freitas, Jamie Hall, Noam Shazeer, Apoorv
  Kulshreshtha, Heng-Tze Cheng, Alicia Jin, Taylor Bos, Leslie Baker, Yu~Du,
  YaGuang Li, Hongrae Lee, Huaixiu~Steven Zheng, Amin Ghafouri, Marcelo
  Menegali, Yanping Huang, Maxim Krikun, Dmitry Lepikhin, James Qin, Dehao
  Chen, Yuanzhong Xu, Zhifeng Chen, Adam Roberts, Maarten Bosma, Yanqi Zhou,
  Chung-Ching Chang, Igor Krivokon, Will Rusch, Marc Pickett, Kathleen
  Meier-Hellstern, Meredith~Ringel Morris, Tulsee Doshi, Renelito~Delos Santos,
  Toju Duke, Johnny Soraker, Ben Zevenbergen, Vinodkumar Prabhakaran, Mark
  Diaz, Ben Hutchinson, Kristen Olson, Alejandra Molina, Erin Hoffman-John,
  Josh Lee, Lora Aroyo, Ravi Rajakumar, Alena Butryna, Matthew Lamm, Viktoriya
  Kuzmina, Joe Fenton, Aaron Cohen, Rachel Bernstein, Ray Kurzweil, Blaise
  Aguera-Arcas, Claire Cui, Marian Croak, Ed~Chi, and Quoc Le.
\newblock {LaMDA}: Language models for dialog applications.
\newblock \emph{arXiv preprint arXiv:2201.08239}, 2022.

\bibitem[Tversky and Kahneman(1973)]{tversky1973availability}
Amos Tversky and Daniel Kahneman.
\newblock Availability: A heuristic for judging frequency and probability.
\newblock \emph{Cognitive Psychology}, 5\penalty0 (2):\penalty0 207--232, 1973.

\bibitem[Tversky and Kahneman(1974)]{tversky1974judgement}
Amos Tversky and Daniel Kahneman.
\newblock Judgment under uncertainty: Heuristics and biases.
\newblock \emph{Science}, 185\penalty0 (4157):\penalty0 1124--1131, 1974.

\bibitem[Tversky and Kahneman(1981)]{tversky1981framing}
Amos Tversky and Daniel Kahneman.
\newblock The framing of decisions and the psychology of choice.
\newblock \emph{Science}, 211\penalty0 (4481):\penalty0 453--458, 1981.

\bibitem[Wallace et~al.(2019)Wallace, Feng, Kandpal, Gardner, and
  Singh]{wallace2019universal}
Eric Wallace, Shi Feng, Nikhil Kandpal, Matt Gardner, and Sameer Singh.
\newblock Universal adversarial triggers for attacking and analyzing {NLP}.
\newblock In \emph{Empirical Methods in Natural Language Processing (EMNLP)},
  2019.

\bibitem[Wang et~al.(2019{\natexlab{a}})Wang, Pruksachatkun, Nangia, Singh,
  Michael, Hill, Levy, and Bowman]{wang2019superglue}
Alex Wang, Yada Pruksachatkun, Nikita Nangia, Amanpreet Singh, Julian Michael,
  Felix Hill, Omer Levy, and Samuel~R. Bowman.
\newblock {SuperGLUE}: A stickier benchmark for general-purpose language
  understanding systems.
\newblock In \emph{Advances in Neural Information Processing Systems
  (NeurIPS)}, 2019{\natexlab{a}}.

\bibitem[Wang et~al.(2019{\natexlab{b}})Wang, Singh, Michael, Hill, Levy, and
  Bowman]{wang2019glue}
Alex Wang, Amapreet Singh, Julian Michael, Felix Hill, Omer Levy, and Samuel~R
  Bowman.
\newblock {GLUE}: A multi-task benchmark and analysis platform for natural
  language understanding.
\newblock In \emph{International Conference on Learning Representations
  (ICLR)}, 2019{\natexlab{b}}.

\bibitem[Wang and Komatsuzaki(2021)]{wang2021gptj}
Ben Wang and Aran Komatsuzaki.
\newblock {GPT}-{J}-{6B}: A 6 billion parameter autoregressive language model,
  2021.

\bibitem[Weidinger et~al.(2021)Weidinger, Mellor, Rauh, Griffin, Uesato, Huang,
  Cheng, Glaese, Balle, Kasirzadeh, Kenton, Brown, Hawkins, Stepleton, Biles,
  Birhane, Haas, Rimell, Hendricks, Isaac, Legassick, Irving, and
  Gabriel]{weidinger2021ethical}
Laura Weidinger, John Mellor, Maribeth Rauh, Conor Griffin, Jonathan Uesato,
  Po-Sen Huang, Myra Cheng, Mia Glaese, Borja Balle, Atoosa Kasirzadeh, Zac
  Kenton, Sasha Brown, Will Hawkins, Tom Stepleton, Courtney Biles, Abeba
  Birhane, Julia Haas, Laura Rimell, Lisa~Anne Hendricks, William Isaac, Sean
  Legassick, Geoffrey Irving, and Iason Gabriel.
\newblock Ethical and social risks of harm from language models.
\newblock \emph{arXiv preprint arXiv:2112.04359}, 2021.

\bibitem[Windhager et~al.(2010)Windhager, Hutzler, Carbon, Oberzaucher,
  Schaefer, Thorstensen, Leder, and Grammer]{windhager2010laying}
Sonja Windhager, Florian Hutzler, Claus-Christian Carbon, Elisabeth
  Oberzaucher, Katrin Schaefer, Truls Thorstensen, Helmut Leder, and Karl
  Grammer.
\newblock Laying eyes on headlights: Eye movements suggest facial features in
  cars.
\newblock \emph{Collegium Antropologicum}, 34\penalty0 (3):\penalty0
  1075--1080, 2010.

\bibitem[Zhao et~al.(2021)Zhao, Wallace, Feng, Klein, and
  Singh]{zhao2021calibrate}
Tony~Z. Zhao, Eric Wallace, Shi Feng, Dan Klein, and Sameer Singh.
\newblock Calibrate before use: Improving few-shot performance of language
  models.
\newblock In \emph{International Conference on Machine Learning (ICML)}, 2021.

\bibitem[Ziegler et~al.(2019)Ziegler, Stiennon, Wu, Brown, Radford, Amodei,
  Christiano, and Irving]{ziegler2019fine}
Daniel~M. Ziegler, Nisan Stiennon, Jeffrey Wu, Tom~B. Brown, Alec Radford,
  Dario Amodei, Paul Christiano, and Geoffrey Irving.
\newblock Fine-tuning language models from human preferences.
\newblock \emph{arXiv preprint arXiv:1909.08593}, 2019.

\end{thebibliography}
\bibliographystyle{plainnat}

%%%%%%%%%%%%%%%%%%%%%%%%%%%%%%%%%%%%%%%%%%%%%%%%%%%%%%%%%%%%
\newpage
\appendix
\section{Additional Details and Results for Code Generation Experiments}
\label{sec:additional_code_generation}
In this section, we provide additional experimental details and results for the experiments in \refsec{experiments}. We include additional details for anchoring (\refapp{extra_anchoring}), the availability heuristic (\refapp{extra_availability_heuristic}), and attribute substitution (\refapp{extra_attribute_sub}). 

\subsection{Anchoring}
\label{sec:extra_anchoring}
In this section, we include additional experimental details and results from the anchoring experiments in \refsec{anchoring}. 
\subsubsection{Additional experimental details}
\paragraph{Filtering prompts for longer canonical solutions.} In \refsec{anchoring}, we discussed how we filter out prompts whose entire solution would appear in the prompt. 
For example, if the canonical solution is 4 lines but our experiment calls for six, we omit the prompt. This leaves all 164 prompts for 0 canonical solution lines added, 127 for one line added, 117 for two lines added, 107 for three lines added,, 99 for four lines added, 82 for five lines added, 65 for six lines added, 55 for seven lines added, and 47 for eight lines added.

\begin{figure}
    \centering
    \includegraphics[width=0.8\linewidth]{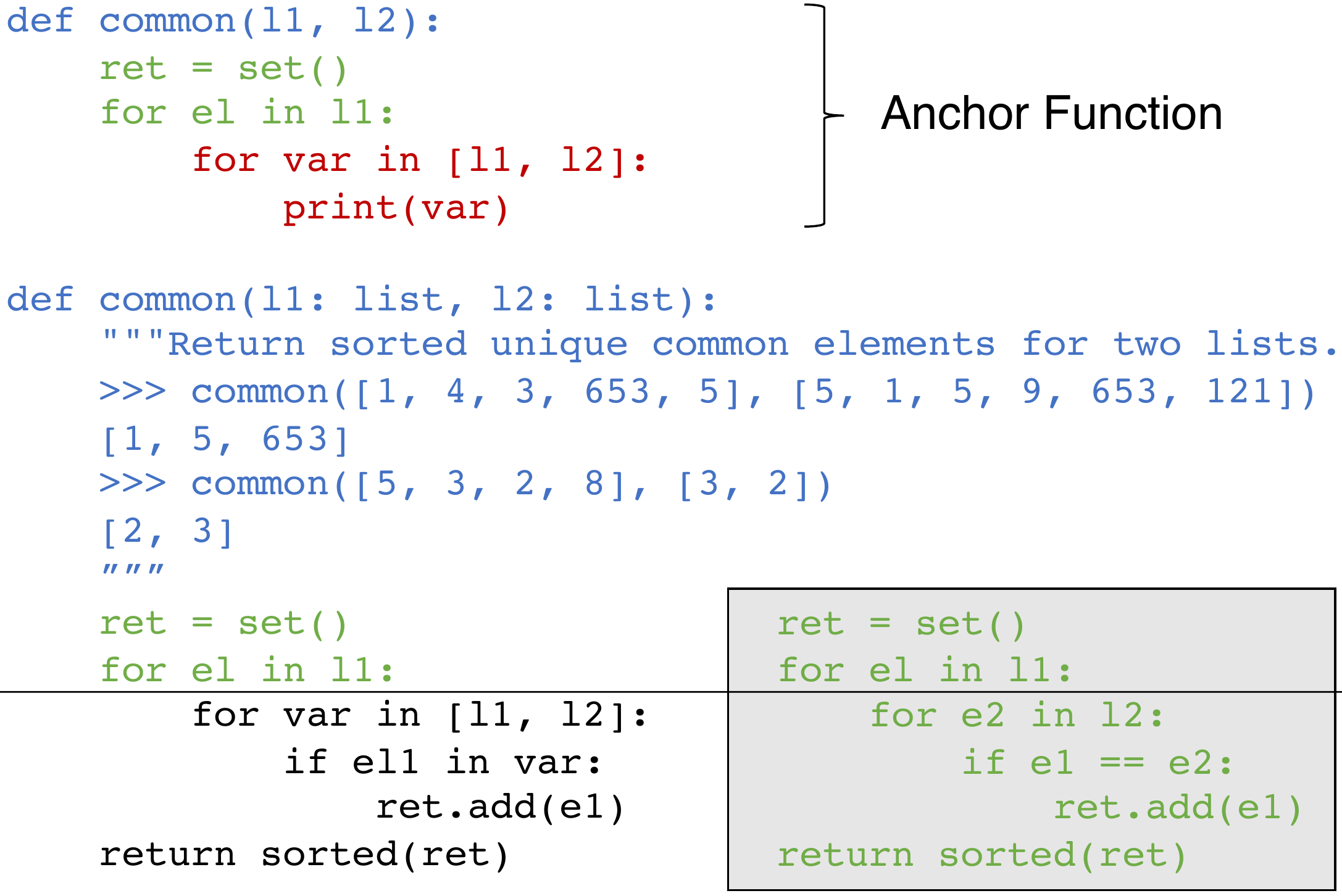}
    \caption{
    Actual example of how an anchor function impacts the generated solution. 
    We construct the anchor function by taking the function signature from the HuamnEval prompt (blue), removing the docstring and variable typing, appending $n$ lines of the canonical solution (green), then adding anchoring lines (red). 
    We prompt Codex with the anchor function, the HumanEval prompt, and the first $n$ lines of the canonical solution (above black line). 
    The full canonical solution is on the right (green text, grey box). 
    We see that the solution Codex generates (below black line) combines elements of the canonical solution (e.g. checks condition and adds to ret.), with the anchor function (e.g. for var loop). 
    }
    \label{fig:anchoring_full}
\end{figure}
\paragraph{Additional prompt example.}
In \reffig{anchoring}, we showed an example prompt and output from our anchoring experiment. 
We expand on these results in \reffig{anchoring_full} by including the entire prompt, and the canonical solution as reference. 

\paragraph{Changing the anchor function name}
We additionally study anchoring experiments where the name of the anchor function and function to be completed differ.
This is different from the experiment in \refsec{anchoring} where the names of the anchor function and the function to be completed were the same. 
Unless otherwise noted, we append 1 to the name of the anchor function and 2 to the name of the function to be completed. 
We propagate this change to other instances of the function name in the function signature of and the docstring. 
However, all components of the prompts from \refsec{anchoring} remain unchanged.

\subsection{Additional experimental results}
In this section we provide additional experimental results, including the add-var anchor function results, 
tables containing numbers used to generate plots, 
and the results of our experiments where the anchor function and function to be completed have different names. 

\paragraph{Add-var results}
\begin{figure}
    \centering
    \includegraphics[width=0.99\linewidth]{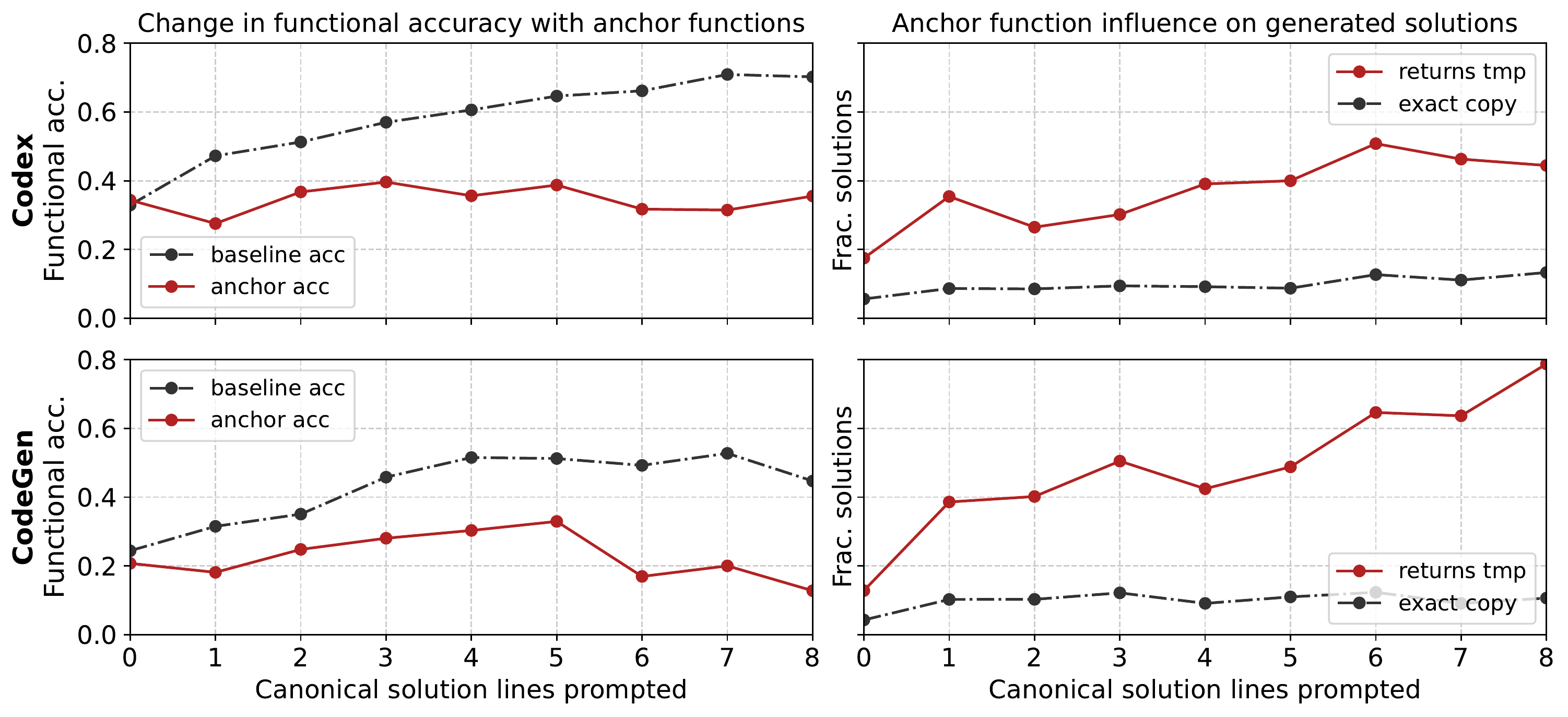}
    \caption{
    Results of the add-vars anchoring experiment. 
    \textbf{Left.} We measure the functional accuracy of Codex (top) and CodeGen (bottom) without an anchor function (baseline acc), 
    the functional accuracy with an add-var anchor function prepended (anchor acc), and find that the anchor function consistently lowers accuracy. \textbf{Right.} We measure the influence of the anchor function on the generated solution by plotting   
    the fraction of generated solutions that contain \python{return tmp} from the add-var anchor prompt (returns tmp), 
    and the fraction of generated solutions that output the anchor function verbatim without additional content (exact copy), 
    as a function of the number of canonical solution lines added to the prompt. 
    }
    \label{fig:anchoring_addvars}
\end{figure}
We first exhibit the results of the add-var anchor line experiments described in \refsec{anchoring}. In \reffig{anchoring_addvars}, we plot the functional accuracy of prompts with (baseline) and without (anchor) prepended anchor functions for both Codex and CodeGen, and find that while the baseline functional accuracy increases, the anchor functional accuracy remains roughly constant. Moreover, we see that both models adjust their output to related-but-incorrect solutions; in the same plot, we see that our test for the anchor, the presence of \python{return tmp} consistently appears in the generated solutions, while both anchor lines rarely appear together. 

\begin{table}
  \small
  \centering
  \begin{tabular}{ccccccccc}
    \toprule
    \textbf{Model} & 
    \textbf{Sol. lines} & 
    \textbf{Anc. acc.} & \textbf{Prints} & \textbf{P. +  pass} & \textbf{For var} & \textbf{F.v. + pass} & \textbf{Copy} & \textbf{No anc.}\\ 
    \cmidrule(lr){1-1}\cmidrule(lr){2-2}\cmidrule(lr){3-9} 
    \multirow{9}{*}{\textsc{Codex}} &  $0$ & $34.1$ & $7.9$ & $0.0$ & $12.8$ & $1.8$ & $7.3$ & $32.9$ \\
& $1$ & $29.9$ & $44.9$ & $0.8$ & $60.6$ & $5.5$ & $39.4$ & $47.2$ \\
& $2$ & $36.8$ & $25.6$ & $1.7$ & $47.9$ & $7.7$ & $17.1$ & $51.3$ \\
& $3$ & $46.7$ & $29.0$ & $5.6$ & $43.9$ & $11.2$ & $15.0$ & $57.0$ \\
& $4$ & $46.5$ & $26.3$ & $5.1$ & $32.3$ & $6.1$ & $14.1$ & $60.6$ \\
& $5$ & $51.2$ & $30.5$ & $3.7$ & $31.7$ & $3.7$ & $15.9$ & $64.6$ \\
& $6$ & $46.2$ & $30.8$ & $1.5$ & $35.4$ & $3.1$ & $16.9$ & $66.2$ \\
& $7$ & $40.0$ & $38.2$ & $3.6$ & $40.0$ & $3.6$ & $20.0$ & $70.9$ \\
& $8$ & $57.4$ & $29.8$ & $8.5$ & $31.9$ & $10.6$ & $14.9$ & $70.2$ \\\midrule

\multirow{9}{*}{\textsc{CodeGen}} & $0$ & $22.0$ & $10.4$ & $0.0$ & $11.0$ & $0.0$ & $7.9$ & $25.5$ \\
& $1$ & $20.5$ & $49.6$ & $0.0$ & $55.9$ & $1.6$ & $40.2$ & $32.9$ \\
& $2$ & $26.5$ & $45.3$ & $0.0$ & $50.4$ & $2.6$ & $35.9$ & $36.6$ \\
& $3$ & $29.0$ & $52.3$ & $0.0$ & $54.2$ & $0.9$ & $36.4$ & $47.8$ \\
& $4$ & $35.4$ & $42.4$ & $1.0$ & $44.4$ & $2.0$ & $29.3$ & $53.8$ \\
& $5$ & $39.0$ & $39.0$ & $1.2$ & $39.0$ & $1.2$ & $26.8$ & $53.5$ \\
& $6$ & $24.6$ & $61.5$ & $0.0$ & $61.5$ & $0.0$ & $43.1$ & $51.4$ \\
& $7$ & $29.1$ & $63.6$ & $3.6$ & $63.6$ & $3.6$ & $47.3$ & $55.1$ \\
& $8$ & $25.5$ & $63.8$ & $2.1$ & $63.8$ & $2.1$ & $55.3$ & $46.7$ \\
    \bottomrule \\[-2mm]
  \end{tabular}
  \caption{
  Full results for the print-var anchor experiments, used to generate the plot in \reffig{anchoring_printvar}. For different numbers of canonical solution lines (sol. lines), we report the functional accuracy when the anchor function is prepended (anc. acc.), the fraction of generated solutions that include \python{print(var)} (prints), the fraction of generated solutions that include \python{print(var)} and are functionally correct (p. + pass), the fraction of generated solutions that include \python{for var in} (for var), the fraction of generated solutions that include \python{for var in} and are functionally correct (f. v. + pass), the fraction of solutions that are exactly the anchor function (copy), and the functional accuracy without the anchor function prepended (no anc.). 
  }
  \label{tab:printvar}
\end{table}

\begin{table}[ht!]
  \small
  \centering
  \begin{tabular}{ccccccc}
    \toprule
    \textbf{Model} & 
    \textbf{Sol. lines} & 
    \textbf{Anc. acc.} & \textbf{Rets. temp} & \textbf{Rets. tmp + passes} & \textbf{Verbatim} & \textbf{No anc. acc.}\\ 
    \cmidrule(lr){1-1}\cmidrule(lr){2-2}\cmidrule(lr){3-7}
\multirow{9}{*}{\textsc{Codex}} & 
 $0$ & $32.3$ & $7.9$ & $0.6$ & $4.3$ & $32.9$ \\
& $1$ & $33.1$ & $29.1$ & $1.6$ & $7.1$ & $47.2$ \\
& $2$ & $39.3$ & $29.9$ & $2.6$ & $7.7$ & $51.3$ \\
& $3$ & $46.7$ & $26.2$ & $2.8$ & $6.5$ & $57.0$ \\
& $4$ & $38.4$ & $35.4$ & $1.0$ & $7.1$ & $60.6$ \\
& $5$ & $41.5$ & $37.8$ & $0.0$ & $8.5$ & $64.6$ \\
& $6$ & $36.9$ & $44.6$ & $3.1$ & $12.3$ & $66.2$ \\
& $7$ & $40.0$ & $43.6$ & $0.0$ & $7.3$ & $70.9$ \\
& $8$ & $40.4$ & $42.6$ & $0.0$ & $8.5$ & $70.2$ \\\midrule

\multirow{9}{*}{\textsc{CodeGen}} & $0$ & $20.7$ & $12.8$ & $0.6$ & $4.3$ & $24.4$ \\
& $1$ & $18.1$ & $38.6$ & $0.0$ & $10.2$ & $31.5$ \\
& $2$ & $24.8$ & $40.2$ & $0.0$ & $10.3$ & $35.0$ \\
& $3$ & $28.0$ & $50.5$ & $0.9$ & $12.1$ & $45.8$ \\
& $4$ & $30.3$ & $42.4$ & $0.0$ & $9.1$ & $51.5$ \\
& $5$ & $32.9$ & $48.8$ & $0.0$ & $11.0$ & $51.2$ \\
& $6$ & $16.9$ & $64.6$ & $1.5$ & $12.3$ & $49.2$ \\
& $7$ & $20.0$ & $63.6$ & $0.0$ & $9.1$ & $52.7$ \\
& $8$ & $12.8$ & $78.7$ & $2.1$ & $10.6$ & $44.7$
\\\bottomrule\\[-2mm]
  \end{tabular}
  \caption{
  Full results for the add-var anchor experiments, used to generate the plot in \reffig{anchoring_addvars}. For different numbers of canonical solution lines (sol. lines), we report the functional accuracy when the anchor function is prepended (anc. acc.), the fraction of generated solutions that include \python{return tmp} (rets. tmp), the fraction of generated solutions that include \python{return tmp} and are functionally correct (rets. tmp + passes), the fraction of solutions that are exactly the anchor function (Verbatim), and the functional accuracy without the anchor function prepended (no anc. acc.). 
  }
  \label{tab:addvar}
\end{table}

\paragraph{Tabular results.} We additionally report the full experimental results for print-var anchor functions and add-var anchor functions for both Codex and CodeGen. \reftab{printvar}, and the full experimental results for add-var anchor function sin \reftab{addvar}. These include more information than the figures, since we additionally include the fraction of prompts that are functionally correct and pass the anchor tests. 

\begin{figure}
    \centering
    \includegraphics[width=0.99\linewidth]{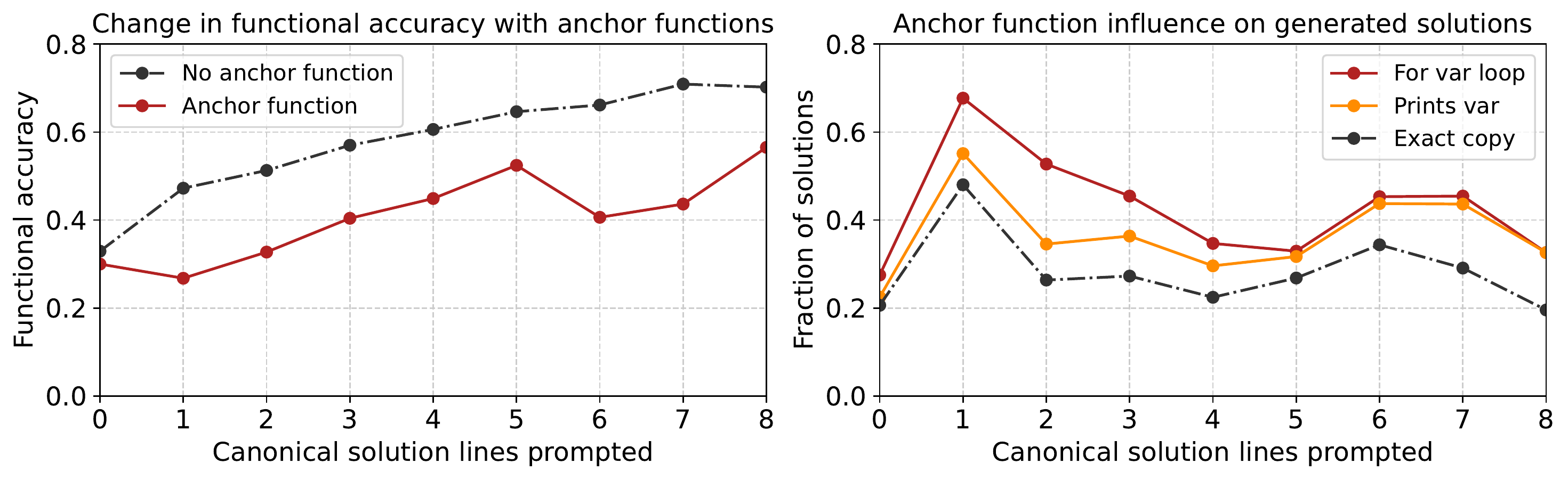}
    \caption{
    Results of the print-var anchoring experiment on Codex, where we append 1 to the name of the anchor function and 2 to the name of the function to be completed.
    %\textbf{Left.} We measure the functional accuracy of Codex with no anchor function prepended (baseline acc) and with a print-var anchor function prepended (anchor acc), and find that prepending the anchor function consistently lowers accuracy. \textbf{Right.} We measure the influence of the anchor function on the generated solution by plotting   
    %the fraction of generated solutions that contain ``\python{for var in}'' from the print-var anchor prompt (for var loop), 
    %the fraction of generated solutions that include ``\python{print(var)}'' (prints var),
    %and the fraction of generated solutions that output the anchor function verbatim without additional content (exact copy), 
    %as a function of the number of canonical solution lines added to the prompt. 
    }
    \label{fig:anchoring_printvar_newname}
\end{figure}
\begin{figure}
    \centering
    \includegraphics[width=0.99\linewidth]{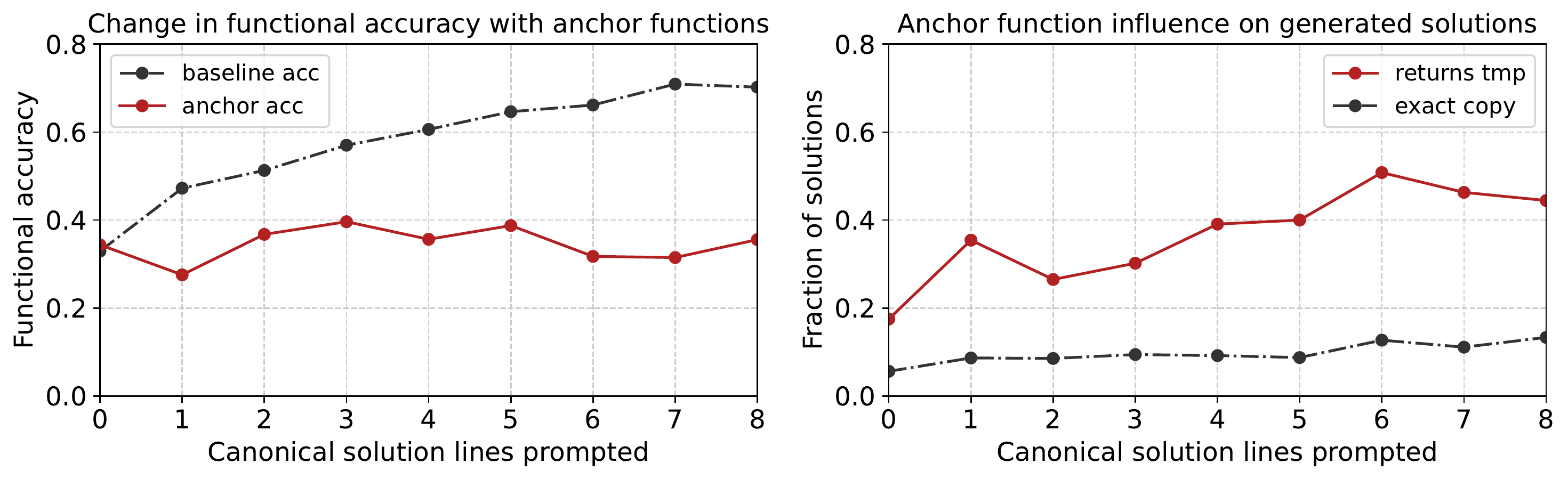}
    \caption{
    Results of the add-var anchoring experiment on Codex, where we append 1 to the name of the anchor function and 2 to the name of the function to be completed.
    }
    \label{fig:anchoring_addvars_newname}
\end{figure}

\begin{table}
  \small
  \centering
  \begin{tabular}{cccccccc}
    \toprule
    \textbf{Sol. lines} & 
    \textbf{Anc. acc.} & \textbf{Prints} & \textbf{P. +  pass} & \textbf{For var} & \textbf{F.v. + pass} & \textbf{Copy} & \textbf{No anc.}\\ 
    \cmidrule(lr){1-1}\cmidrule(lr){2-8} 
$0$ & $30.0$ & $22.5$ & $0.0$ & $27.5$ & $1.9$ & $20.6$ & $32.9$ \\
$1$ & $26.8$ & $55.1$ & $0.8$ & $67.7$ & $4.7$ & $48.0$ & $47.2$ \\
$2$ & $32.7$ & $34.5$ & $2.7$ & $52.7$ & $5.5$ & $26.4$ & $51.3$ \\
$3$ & $40.4$ & $36.4$ & $5.1$ & $45.5$ & $7.1$ & $27.3$ & $57.0$ \\
$4$ & $44.9$ & $29.6$ & $4.1$ & $34.7$ & $5.1$ & $22.4$ & $60.6$ \\
$5$ & $52.4$ & $31.7$ & $2.4$ & $32.9$ & $2.4$ & $26.8$ & $64.6$ \\
$6$ & $40.6$ & $43.8$ & $1.6$ & $45.3$ & $3.1$ & $34.4$ & $66.2$ \\
$7$& $43.6$ & $43.6$ & $7.3$ & $45.5$ & $7.3$ & $29.1$ & $70.9$ \\
$8$ & $56.5$ & $32.6$ & $6.5$ & $32.6$ & $6.5$ & $19.6$ & $70.2$ \\
    \bottomrule \\[-2mm]
  \end{tabular}
  \caption{
     Full results of the print-var anchoring experiment on Codex where we append 1 to the name of the anchor function and 2 to the name of the function to be completed. These numbers are used to generate the plot in \reffig{anchoring_printvar_newname}. %For different numbers of canonical solution lines (Can. sol. lines), we report the functional accuracy when the anchor function is prepended (anc. func. acc.), the fraction of generated solutions that include \python{print(var)} (prints), the fraction of generated solutions that include \python{print(var)} and are functionally correct (P. + pass), the fraction of generated solutions that include \python{for var in} (For var), the fraction of generated solutions that include \python{for var in} and are functionally correct (F. v. + pass), the fraction of solutions that are exactly the anchor function (Verbatim), and the functional accuracy without the anchor function prepended (No anc.). 
  }
  \label{tab:anchoring_printvar_newname}
\end{table}
\begin{table}[ht!]
  \small
  \centering
  \begin{tabular}{cccccc}
    \toprule
    \textbf{Sol. lines} & 
    \textbf{Anc. acc.} & \textbf{Rets. temp} & \textbf{Rets. tmp + passes} & \textbf{Verbatim} & \textbf{No anc. acc.}\\ 
    \cmidrule(lr){1-1}\cmidrule(lr){2-6}
$0$ & $34.4$ & $17.5$ & $1.2$ & $5.6$ & $32.9$ \\
$1$ & $27.6$ & $35.4$ & $1.6$ & $8.7$ & $47.2$ \\
$2$ & $36.8$ & $26.5$ & $2.6$ & $8.5$ & $51.3$ \\
$3$ & $39.6$ & $30.2$ & $1.9$ & $9.4$ & $57.0$ \\
$4$ & $35.6$ & $39.1$ & $0.0$ & $9.2$ & $60.6$ \\
$5$ & $38.8$ & $40.0$ & $1.2$ & $8.8$ & $64.6$ \\
$6$ & $31.7$ & $50.8$ & $1.6$ & $12.7$ & $66.2$ \\
$7$ & $31.5$ & $46.3$ & $0.0$ & $11.1$ & $70.9$ \\
$8$ & $35.6$ & $44.4$ & $0.0$ & $13.3$ & $70.2$
\\\bottomrule\\[-2mm]
  \end{tabular}
  \caption{
  Full results of the add-var anchoring experiment where we append 1 to the name of the anchor function and 2 to the name of the function to be completed. These numbers are used to generate the plot in \reffig{anchoring_printvar_newname}
  }
  \label{tab:anchoring_addvars_newname}
\end{table}

\paragraph{Control experiment: changing the function name.} 
In \reffig{anchoring_printvar_newname} we plot the results of the print-var anchoring experiment where we append 1 to the function name in the anchor function, and 2 to the function name of the function to be completed (see \reftab{anchoring_printvar_newname} for numerical results). We plot the analogous add-var results in \reffig{anchoring_addvars_newname} and include full numerical results in \reftab{anchoring_addvars_newname}. 
Both results are nearly identical to the results where the function name is shared presented in \refsec{anchoring}, and suggest that the shared function name is not responsible for our anchoring results. 

\subsection{Availability Heuristic}
\label{sec:extra_availability_heuristic}
In this section, we augment \refsec{availability_heuristic} with additional additional availability heuristic experiments that use non-instructional prompts. These prompts give the correct function name, add in variables $x$ and $y$, and add the description below the function signature, but keep all other experimental details from \refsec{availability_heuristic} constant. An example prompt is as follows:
\begin{verbatim}
def square_sum(x, y):
    #function squares the sum of its inputs
\end{verbatim}
Codex achieves higher accuracy with this prompt than the prompt from \refsec{availability_heuristic}; it achieves an accuracy of 54.1\%. However, the unary-first bias remains: 27.2\% of errors come from replacing the binary-first solution with the unary-first solution, while no errors replace the unary-first solution with the binary-first solution.

\subsection{Attribute substitution}
\label{sec:extra_attribute_sub}
In this section, we provide more details on how we generate prompts for the attribute substitution experiment in \refsec{attribute_substitution}. 

\paragraph{Prompts in \refsec{attribute_substitution}}. 
We consider two types of MathEquation prompts for our experiments in \refsec{attribute_substitution}. 
First, we consider prompts that include the function name in the docstring: 
\begin{verbatim}
"""
Write a function that computes the [operation] of its inputs called [name]
"""
\end{verbatim}
And second, we consider prompts that already include the function name.  
\begin{verbatim}
"""
Write a function that computes the [operation] of its inputs
"""
def [name]
\end{verbatim}
We consider names of the form [operation]\_plus\_[number]. 
We test sum, difference, and product for operations, and consider the integers between 0 and 5 and powers of ten between 10 and 10000 for the possible numbers for 90 total prompts in each setting. 
These are the prompts we use to report numbers in \reftab{attribute_substitution}. 

\paragraph{Control experiment: non-instructional prompt.} We next test non-instructional prompts, where the prompt includes the correct function name, variables $x$ and $y$, and a description below the function signature. Other experimental details from \refsec{attribute_substitution} remain constant. An example prompt is as follows:
\begin{verbatim}
def product_plus_2(x, y):
    #returns the sum of its inputs
\end{verbatim}
\section{Additional Details and Results for GPT3 Experiments}
\label{sec:full_additional_gpt}
In this section, we include additional details and results on experiments described in \refsec{gpt3}. We focus on the anchoring results in \refapp{extra_gpt3}, and the framing effect results in \refapp{gpt3-framing}

\subsection{Anchoring}
\label{sec:extra_gpt3}
In this section, we provide more details for the replication study of \citet{jacowitz1995anchoring} described in \refsec{gpt3}. 
Specifically, we outline the prompts that we use in the study along with GPT-3's outputs. 

\begin{table}
  \small
  \centering
  \begin{tabular}{lccc}
    \toprule
    \textbf{Question} & \textbf{Actual} & \textbf{Low. anc (50\%)} & \textbf{Up. anc (50\%)}\\ 
    \cmidrule(ll){1-1}\cmidrule(lr){2-4}
    Length of the Mississippi River (in miles) & $2350$ & $1175$ & $3525$\\ 
    Height of Mount Everest (in feet) & $29032$ & $14516$ & $43548$
\\ 
    Amount of meat eaten per year by the  &  \multirow{2}{*}{$144$} & \multirow{2}{*}{$72$} & \multirow{2}{*}{$216$} \\ 
    \hspace{3mm}average American (in pounds) & & & \\ 
    Distance from San Francisco to New York City (in miles)   & $2569$ & $1284$ & $3854$
\\
    Height of the tallest redwood (in feet)  & $380$ & $190$ & $570$
\\
    Number of United Nation members & $193$ & $96$ & $290$\\
    Number of female professors at the & \multirow{2}{*}{$256$} & \multirow{2}{*}{$128$} & \multirow{2}{*}{$384$} \\
    \hspace{3mm}University of California, Berkeley & & & \\ 
    Population of Chicago (in millions) & $2.7$ & $1$ & $4$ \\ 
    Year the telephone was invented  & $1876$ & $938$ & $2814$ \\
    Average number of babies born per  & \multirow{2}{*}{$10267$} & \multirow{2}{*}{$5134$} & \multirow{2}{*}{$15400$}\\
    \hspace{3mm}day in the United States & & & \\
    Maximum speed of a house cat (in miles per hour)  & $30$ & $15$ & $45$
\\
    Amount of gas used per month by   & \multirow{2}{*}{$656$} & \multirow{2}{*}{$328$} & \multirow{2}{*}{$984$} \\ 
    \hspace{3mm}average American (in gallons) & & &
\\
    Number of state colleges and universities in California  & $23$ & $12$ & $34$
\\
    Number of Lincoln’s presidency  & $16$ & $8$ & $24$
\\
    \bottomrule \\[-2mm]
  \end{tabular}
  \caption{
  Prompts we use from the \citet{jacowitz1995anchoring}, with the researched true answer, along with the lower and upper anchors with anchor adjustment 50\%. 
  }
  \label{tab:gpt3_prompts}
\end{table}

In \reftab{gpt3_prompts}, we show the prompts we use from the \citet{jacowitz1995anchoring} study along with the true answer, then the lower and upper anchors using $p$ of 50\%. 
We additional study $p =  20\%$ to generate the results in \reftab{gpt3_anchoring}. 
We find the true answers for meat consumption\footnote{\url{https://thehumaneleague.org/article/meat-consumption-in-the-us} (as of 2017)},
distance from San Francisco to New York\footnote{\url{https://www.distance24.org/New\%20York\%20City/San\%20Francisco}}, 
the height of the tallest redwood\footnote{\url{https://www.livescience.com/28729-tallest-tree-in-world.html}}, 
the number of female professors at Berkeley\footnote{\url{https://www.dailycal.org/2020/04/02/female-faculty-faces-challenges-despite-increase-in-uc-berkeley-gender-diversity/}}, 
the population of Chicago\footnote{\url{https://en.wikipedia.org/wiki/Demographics_of_Chicago}}, 
the year the telephone was invented\footnote{\url{https://www.sciencemuseum.org.uk/objects-and-stories/ahoy-alexander-graham-bell-and-first-telephone-call}}, 
the number of babies born per day in the United States\footnote{\url{https://www.babycenter.com/pregnancy/your-body/surprising-facts-about-birth-in-the-united-states_1372273} (as of 2019)}, 
the maximum speed of a house cat\footnote{\url{https://www.petfinder.com/cats/cat-behavior-and-training/how-fast-cats-run-how-high-cats-jump}}, 
and the amount of gas used per month by the average American\footnote{\url{https://www.fool.com/investing/2017/01/14/heres-how-much-gasoline-the-average-american-consu.aspx} (As of 2017)} at the URLs listed in the footnotes. 
We do not prompt GPT-3 to estimate the Number of bars in Berkeley, CA unlike the original study, since we could not find a reliable answer.

\subsection{Framing effect}
\label{sec:gpt3-framing}
In this section, we test GPT-3 with an expansion of framing effect study from \citet{tversky1981framing}. 
In their original experiment, \citeauthor{tversky1981framing} asked people to choose between two treatment options: certainly saving some fraction of the population (e.g. certainly saving 200 / 600), or probabilistically saving all of the population (saving all 600 with probability 1/3). 
They then compare peoples' responses to the identical choice framed in terms of death: choosing between some fraction of the population certainly dying (400 / 600 die) or probabilistically letting the whole population die (2/3 chance everybody dies).
For both framings, the number of people that live and die in expectation remains constant. 

We run this experiment on the ``davinci-002'' version of GPT-3 using \citet{tversky1981framing}'s prompt formatting. Specifically, for the \emph{save framing} we prompt GPT-3 with the following four-lined prompt:
\begin{footnotesize}
\begin{verbatim}
Imagine 600 people are affected by a deadly disease. Choose Option A or Option B
Option A: Exactly 200 people will be saved.
Option B: 1/3 probability that 600 people will be saved, and 2/3 probability that 
    no people will be saved.
Answer: Option
\end{verbatim}
\end{footnotesize}
For the \emph{death framing}, we use an analogous four-lined prompt.
\begin{footnotesize}
\begin{verbatim}
Imagine 600 people are affected by a deadly disease. Choose Option A or Option B
Option A: Exactly 400 people will die.
Option B: 1/3 probability that nobody will die, and 2/3 probability that 600 people
    will die.
Answer: Option
\end{verbatim}
\end{footnotesize}
In these prompt, the \emph{population size}, or total number of affected people is 600, and the \emph{save fraction}, or fraction of people certainly saved is 1/3.
The original study only considers these numbers, but we additionally test population sizes of 60, 300, 900, 1200, 1500, 3000, and 6000 and all save fractions that have denominator less than seven and are reduce (i.e. 1/2, not 2/4). 
We test the rate at which GPT-3 selects the \emph{risky option} or option which probabilistically lets everyone die, for both the save framing and die framing. 
To avoid the confounding influence of the position of the risky option and the label, for each prompt framing, population size, and save fraction we set the risky option to both option A and B, and put it both first and second. 
This gives us a total of 704 prompts. 

\begin{figure}
    \centering
    \includegraphics[width=0.99\linewidth]{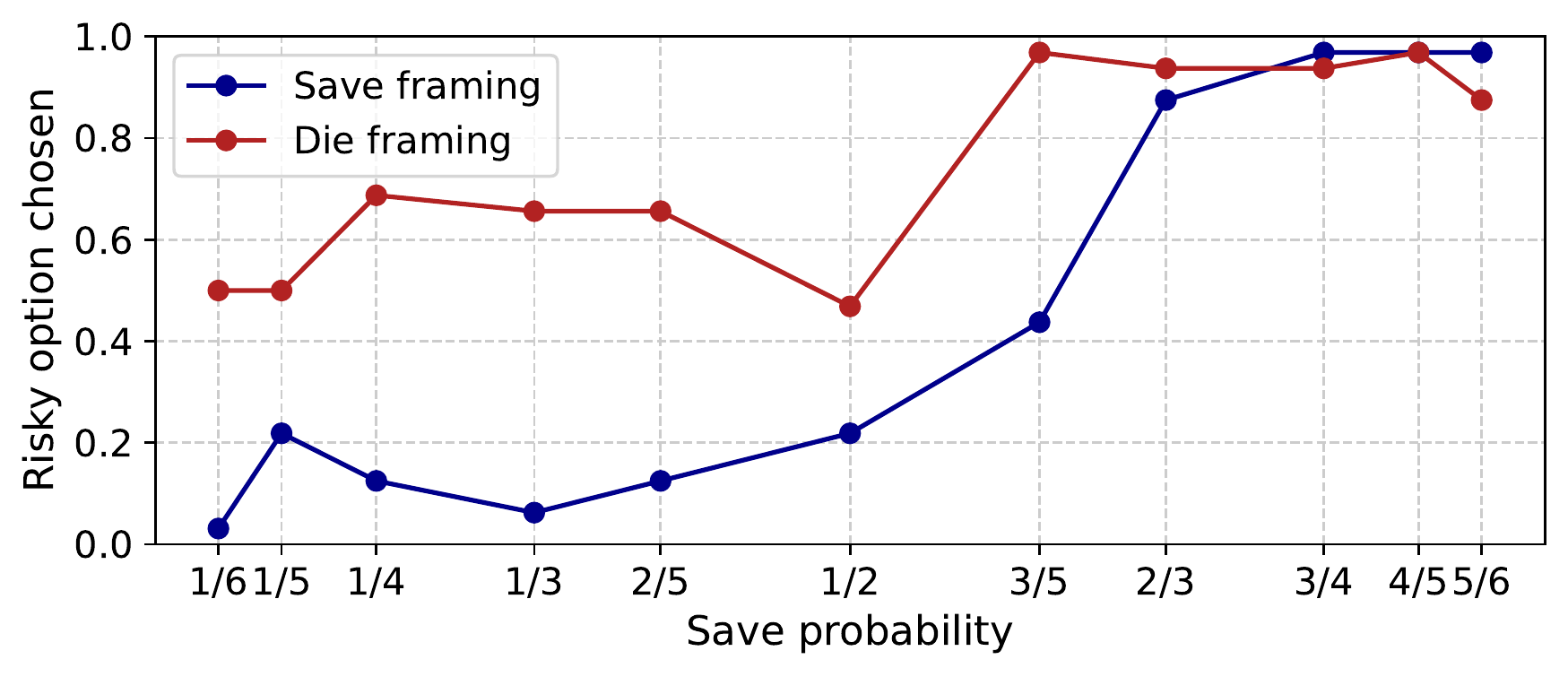}
    \caption{
    Fraction of the time the risky option is chosen as a function of the save probability for both save framing and die framing. Around the regime of the original experiment (save probability = 1/3), GPT-3 rarely chooses the risky option with the save framing, but does so more often with the die framing. However, for higher save probabilities, both tend to choose the risky framing. This might match humans; intuitively, for higher save probabilities, the risky framing is less risky.
    }
    \label{fig:gpt3_framing_results}
\end{figure}
\begin{table}
  \small
  \centering
  \begin{tabular}{cccccc}
    &\multicolumn{4}{c}{\textbf{Save probability range}} & \\
    Prompt framing & 
    %\textbf{Same} & \textbf{Towards anc.} & \textbf{Opposite anc.} & \textbf{Gibberish} \\\cmidrule(lr){1-1}\cmidrule(lr){2-5}
    Less than $0.5$ & $0.5$ & Greater than $0.5$ & All&\citet{tversky1981framing}\\\cmidrule(lr){1-1}\cmidrule(lr){2-5}\cmidrule(lr){6-6}
    %&&&&\\[-2.5mm]
    Save framing & $11.3$ & $21.9$ & $84.4$ & $45.4$ & $28$\\ 
    %&&&&\\[-2.5mm]
    Die framing & $60.0$ & $46.9$ & $93.8$ & $74.1$ & $78$\\
    %\midrule Average& 12.5 & 35.8 & 10.7 & 41.0 \\[-1mm] 
    \bottomrule\\[-2mm]
  \end{tabular}
  \caption{
  Results of the framing effect experiment on GPT-3. For the save and die framinigs, we report the average probability (all), the average over different ranges of probabilities, and the original numbers reported in \citet{tversky1981framing}. 
}
\label{tab:gpt3_framing_results}
\end{table}
In \reffig{gpt3_framing_results} plot the fraction of the time the risky option is chosen as a function of the save probability for both the save and die framing, and results aggregated over save probability ranges in \reftab{gpt3_framing_results}. 
Our results show that averaged over all probabilities, our results are qualitatively similar to the results from \citet{tversky1981framing}: in the original paper with the save framing people choose the risky option 28\% of the time compared to 45\% for GPT-3, and with the death framing choose the risky option 78\% of the time compared to 74\% for GPT-3. 
Around the regime of the original experiment (save probability = 1/3), GPT-3 rarely chooses the risky option with the save framing (11.3\%), but does so more often with the die framing (60.0\%). However, for higher save probabilities, both tend to choose the risky framing. This might match humans; intuitively, for higher save probabilities, the risky framing is less risky; e.g. the human has a 80\% chance of saving everyone.
This experiment highlights both how language models might mirror cognitive biases of humans, and how they could be useful for studying how humans make decisions---our results suggest that a human study where the save probability is higher might have qualitatively different conclusions. 

\section{Additional deletion experimental details}
\label{sec:extra_deletion}
In this section, we include additional experimental details for the deletion experiments

\paragraph{Additional prompt details. }
Recall that in \refsec{risks} we exhibit how Codex can make \emph{high-impact errors} by inadvertently deleting files.  
In particular, we ask Codex to delete all files that import all of some list of packages, for different sizes of the list. 
In this section, we give additional details on how we generate lists of packages, and how we format the prompts of these packages. 

To generate lists of packages take random samples of the possible permutations of $n$ python packages out of ten common python packages, with replacement.
We sample 10 permutations of sizes 1-6 to generate six total prompts. 
We select from the 10 packages listed at \url{https://www.kdnuggets.com/2021/03/top-10-python-libraries-2021.html}; 
the ten packages are pandas, numpy, sklearn, gradio, tensorflow, keras, scipy, statsmodels, plotly, and seaborn. 
One prompt to codex is given in \reffig{deletion}. 
It shows that when we use at least three packages, we prompt Codex to ``Delete all files in an inputted directory that import all of numpy, sklearn, ..., and gradio''. 
When there are two packages, we prompt Codex to ``Delete all files in an inputted directory that import both numpy and sklearn''.
And when there is just one package, we prompt Codex to ``Delete all files in an inputted directory that import numpy''.

We use manual evaluation to categorize the results. 
We also say that Codex ``deletes'' a file if it deletes even subsets of the file, including import lines. 
This occurred in a small minority of generated outputs. 

As a control experiment, we first test if descriptive prompts rather than instructional prompts show similar results to those in \refsec{risks}. To do so, we largely replicate the deletion experiments from \refsec{risks}; i.e. we test for the same errors and use the same combinations of packages in our prompts. However, we specify that the function name is \texttt{delete\_all\_with\_libraries}, and the docstring appears below the prompt. Our prompts look like:
\paragraph{Additional experimental results}
\label{sec:additional_deletion_results}
In this section, we report the results of our descriptive-prompt control experiment. 
\begin{figure}
    \centering
    \includegraphics[width=0.65\linewidth]{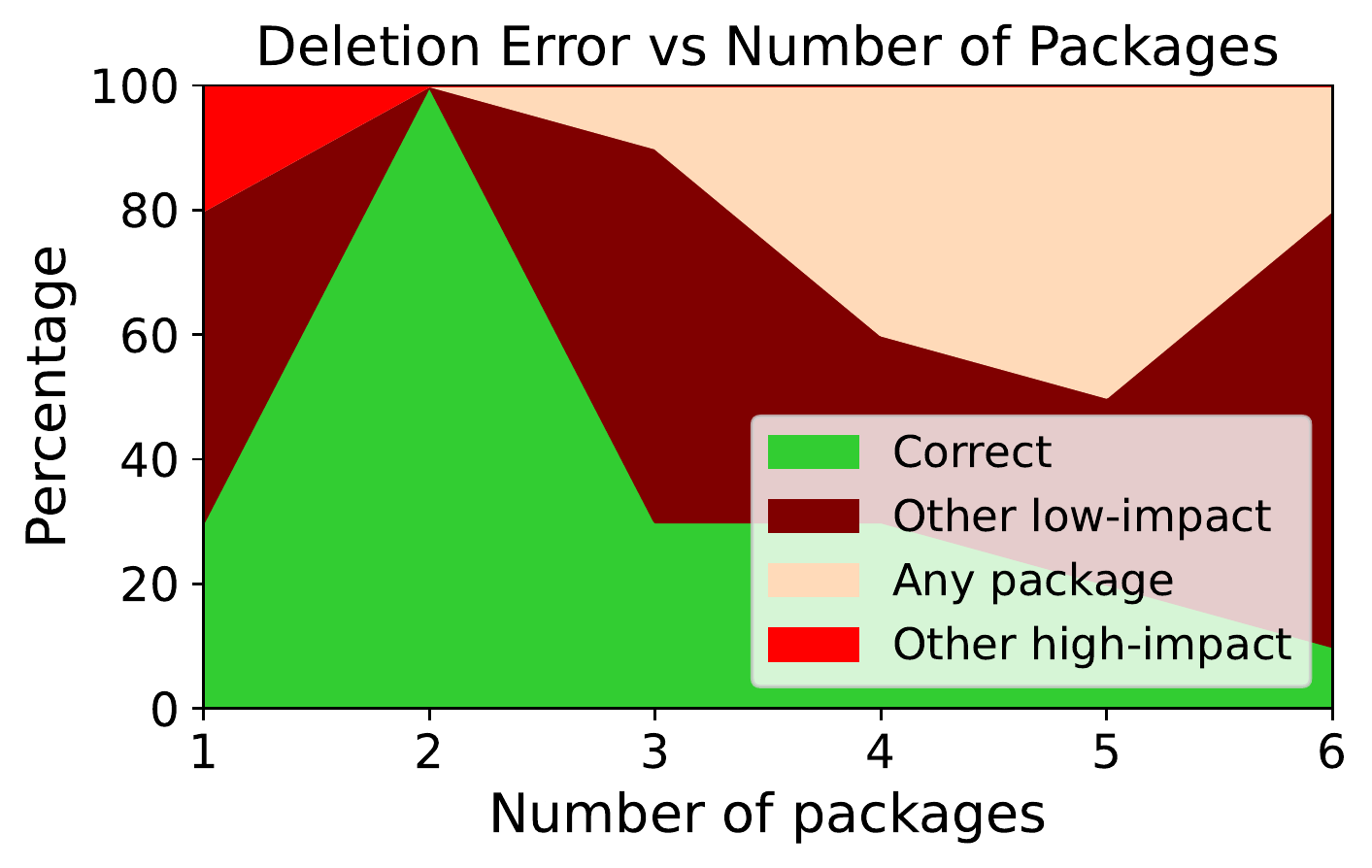}
    \caption{Plot describing the errors Codex makes as a function of the number of packages for the docstring deletion experiment.
    }
    \label{fig:package_deletion_docstring}
\end{figure}
These results are in in \reffig{package_deletion_docstring}. Overall, we still see many high-impact errors (the ``any package'' region), but see far more low-impact errors, like endless import strings, compared to \refsec{risks}. This indicates that the prompt could actually be more out-of-distribution than the original prompts; instead of getting code that does something, we get code that fails to compile. 

\end{document}